\documentclass[letterpaper]{article} 
\usepackage[preprint]{aaai2027}  
\usepackage[hyphens]{url}  
\usepackage{graphicx} 
\usepackage{natbib}  
\usepackage{caption} 
\usepackage{amsmath,amsfonts,bm}

\usepackage{booktabs}
\usepackage[table]{xcolor}
\usepackage{multirow}
\usepackage{pifont}
\usepackage{docmute}
\usepackage{tcolorbox}
\tcbuselibrary{most}
\definecolor{oursrow}{RGB}{221,240,250}
\newcommand{\cmark}{\ding{51}}
\newcommand{\xmark}{\ding{55}}
\tcbset{
  aibox/.style={
    width=\textwidth,
    top=10pt,
    colback=white,
    colframe=black,
    colbacktitle=black,
    enhanced,
    center,
    attach boxed title to top left={yshift=-0.1in,xshift=0.15in},
    boxed title style={boxrule=0pt,colframe=white},
  }
}
\newtcolorbox{AIbox}[2][]{aibox,title=#2,#1}
\newcommand{\sampleimage}[1]{\includegraphics[width=0.30\textwidth,height=0.17\textheight,keepaspectratio]{#1}}
\newcommand{\samplehalfrow}{\rule[-0.0425\textheight]{0pt}{0.085\textheight}}

\title{Aero Realtime: Fully Aligned Input-Output Streams for Low-Latency Streaming Multimodal Generation}
\author{Kaichen Zhang\textsuperscript{1,2}, Wei Huang\textsuperscript{1},
Keming Wu\textsuperscript{2,3}, Bo Li\textsuperscript{2},
Xiaojuan Qi\textsuperscript{1}\corresponding}
\affiliations{\textsuperscript{1}The University of Hong Kong, \textsuperscript{2}LMMs-Lab, \textsuperscript{3}Tsinghua University\\
kaichenzhang@connect.hku.hk, xjqi@eee.hku.hk\\
Project page: \url{https://kcz358.github.io/aero-realtime/}}

\begin{document}

\maketitle

\begin{abstract}
Existing streaming multimodal models process observations incrementally but still follow a turn-based prefill-then-decode pattern, making them non-duplex: new observations cannot naturally enter an active generation stream. Proactive alternatives use micro-turn polling or external response gates, which fragment continuous interaction, decouple response timing from language generation, and complicate KV-cache-friendly serving.
We introduce Aero Realtime, a 4B streaming multimodal model with a duplex architecture for realtime generation. Aero Realtime aligns video, audio, and textual output on a shared temporal grid, where each approximately 80-ms audio slot predicts either a lexical token or a silence token. This allows input and output to advance together, enabling one autoregressive objective to learn both when to respond and what to generate. During inference, Aero Realtime appends only the newest multimodal slot, carries forward the previous output state, and reuses the KV cache for efficient incremental execution.
We further provide a complete training and serving recipe, including realtime QA construction, slot-aligned supervision, hardware-aware distributed training, and resumable inference. On four NVIDIA A6000 workstation GPUs, Aero Realtime maintains 84-ms median and 173-ms P95 processing lag over 20 minutes of a continuously streamed video, remaining within 200~ms of the source timeline. These results demonstrate the feasibility of fully aligned input-output modeling for duplex, proactive, and hardware-aligned multimodal interaction.

\end{abstract}

\section{Introduction}
\label{sec:introduction}


Interactive multimodal intelligence requires a model to continuously perceive the world while responding at the right moment. In realistic scenarios, visual and auditory observations arrive as uninterrupted streams, and the model must decide not only what to say but also when to say it. Recent streaming video-language models have made important progress by processing incoming frames incrementally~\citep{Chen2024VideoLLMOnline,Chen2025Livecc} and maintaining bounded memory over long-running videos~\citep{Xu2025Streamingvlm}. However, \emph{streaming perception} alone is insufficient for realtime interaction. A truly realtime multimodal model should be able to receive new observations while generating a response, remain silent when no response is needed, and execute efficiently under the prefill–decode and KV-cache mechanisms used by modern inference engines.

\begin{figure*}[t]
    \centering
    \includegraphics[width=0.96\textwidth]{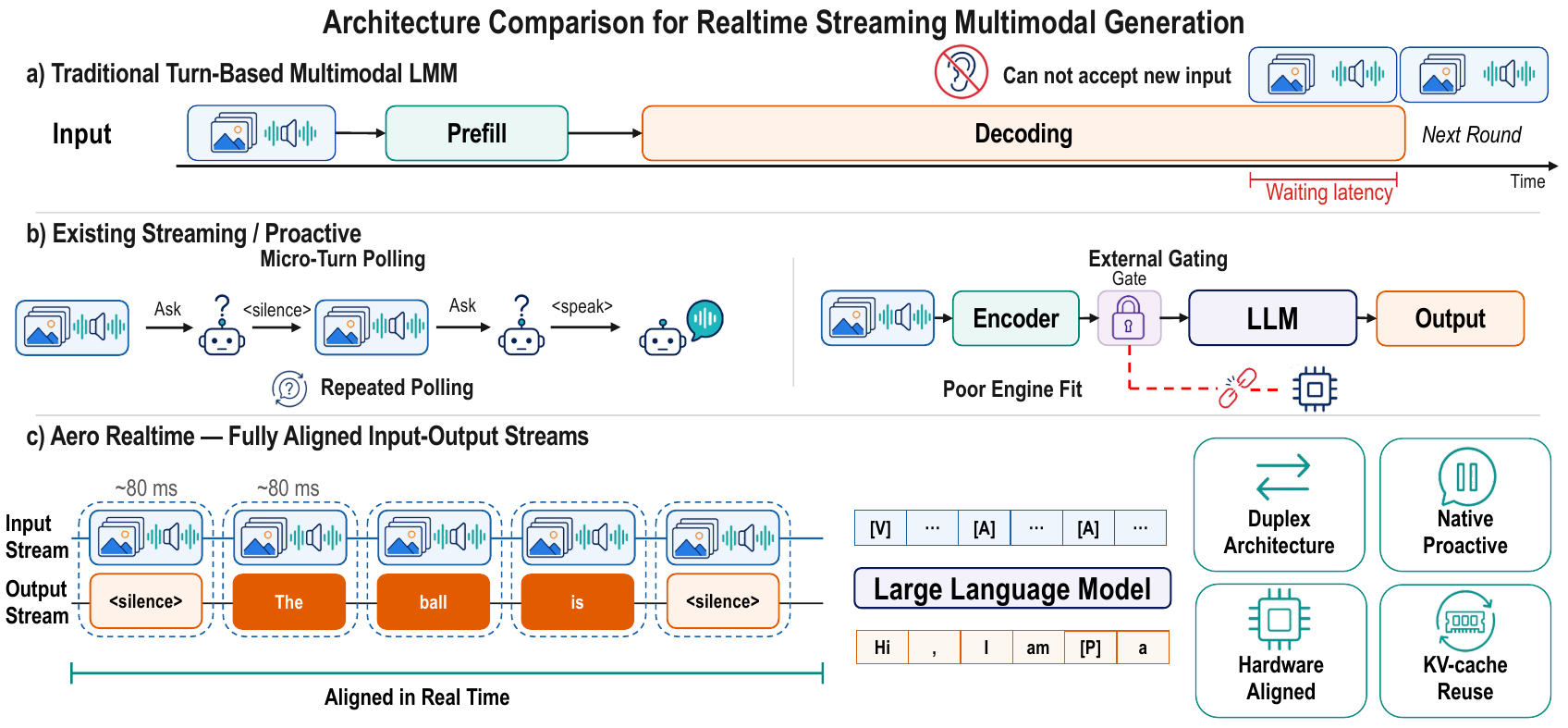}
    \caption{Realtime multimodal architectures: (a) non-duplex turn-based generation, (b) micro-turn polling and external gating, and (c) Aero Realtime's aligned input-output stream.}
    \label{fig:interaction-mechanism}
\end{figure*}

Existing multimodal language models are still largely inherited from turn-based interaction, as illustrated in Figure~\ref{fig:interaction-mechanism}(a). They typically serialize visual, audio, and textual inputs into a multimodal prompt, perform a prefill step to encode this prompt, and then generate response tokens through autoregressive decoding. This input-prefix/output-suffix formulation works well for offline visual question answering and conventional instruction following, but it creates a fundamental mismatch with continuous streaming. Once decoding begins, newly arriving observations cannot naturally enter the active generation sequence. As a result, input and output advance in separate phases rather than on the same temporal clock, creating a non-duplex architecture in which the model cannot truly listen while speaking.


Recent proactive and streaming systems attempt to reduce this mismatch through the two strategies illustrated in Figure~\ref{fig:interaction-mechanism}(b). One common strategy~\citep{Chen2025Livecc,Xu2025Streamingvlm} divides the input stream into short micro-turns and repeatedly queries the model or a controller to determine whether it should respond. This approach improves temporal granularity, but it does not remove the turn boundary; the model still alternates between observation, decision, and generation. Another strategy introduces explicit response-control modules, such as decision heads~\citep{Chen2024VideoLLMOnline}, activation gates~\citep{Qian2025Dispider}, or separated perception-decision-reaction components~\citep{Wang2025Streambridge}. These modules can help determine whether a response is useful, but response timing is no longer naturally learned as part of the language-generation objective. Moreover, external gating complicates deployment because it does not directly align with the prefill-decode execution path and KV-cache reuse optimized in current inference systems.

The central challenge is therefore to unify continuous perception, response timing, and lexical generation without breaking efficient incremental inference. A realtime architecture must admit new observations during generation, model silence and speech under one objective, and preserve KV-cache reuse, as summarized in Table~\ref{tab:interaction-properties}.


\begin{table}[t]
\centering
\small
\setlength{\tabcolsep}{3pt}
\resizebox{\columnwidth}{!}{%
\begin{tabular}{lcccc}
\toprule
Paradigm & Continuous Input & Listen While Speaking & Native Proactive & Engine Aligned \\
\midrule
Turn-Based & \xmark & \xmark & \xmark & \cmark \\
Micro-Turn Polling & \cmark & \xmark & Partial & Partial \\
External Gating & \cmark & Partial & \xmark & \xmark \\
\rowcolor{oursrow} Aero Realtime & \cmark & \cmark & \cmark & \cmark \\
\bottomrule
\end{tabular}
}
\caption{Comparison of interaction paradigms.}
\label{tab:interaction-properties}
\end{table}
We introduce \textbf{Aero Realtime}, a 4B streaming multimodal model for realtime generation, illustrated in Figure~\ref{fig:interaction-mechanism}(c). Aero Realtime places video, audio, and textual output on a shared temporal grid. Each approximately 80-ms audio slot is paired with one output slot containing either a lexical token or a silence token, allowing input and output to advance together. The same language-model head therefore learns both \emph{when} to respond and \emph{what} to generate under one autoregressive objective.

Making this formulation practical requires new supervision, training, and serving mechanisms. First, we develop a unified slot-aligned representation that maps temporally annotated responses onto their causal audio timeline and converts conventional video QA by appending silent audio slots, allowing heterogeneous data to train the same duplex interface. Second, we design modality-aware three-level parallelism: visual frames are distributed according to patch workload, while padding-free audio and packed multimodal language sequences use sequence parallelism along their respective sequence dimensions. Third, we introduce cache-valid delta inference, which appends only the newest multimodal slot, carries the sampled token into the next audio-text state, and preserves the computed prefix across continuous updates without repeated prefill.
\begin{figure*}[t]
    \centering
    \includegraphics[width=0.9\textwidth]{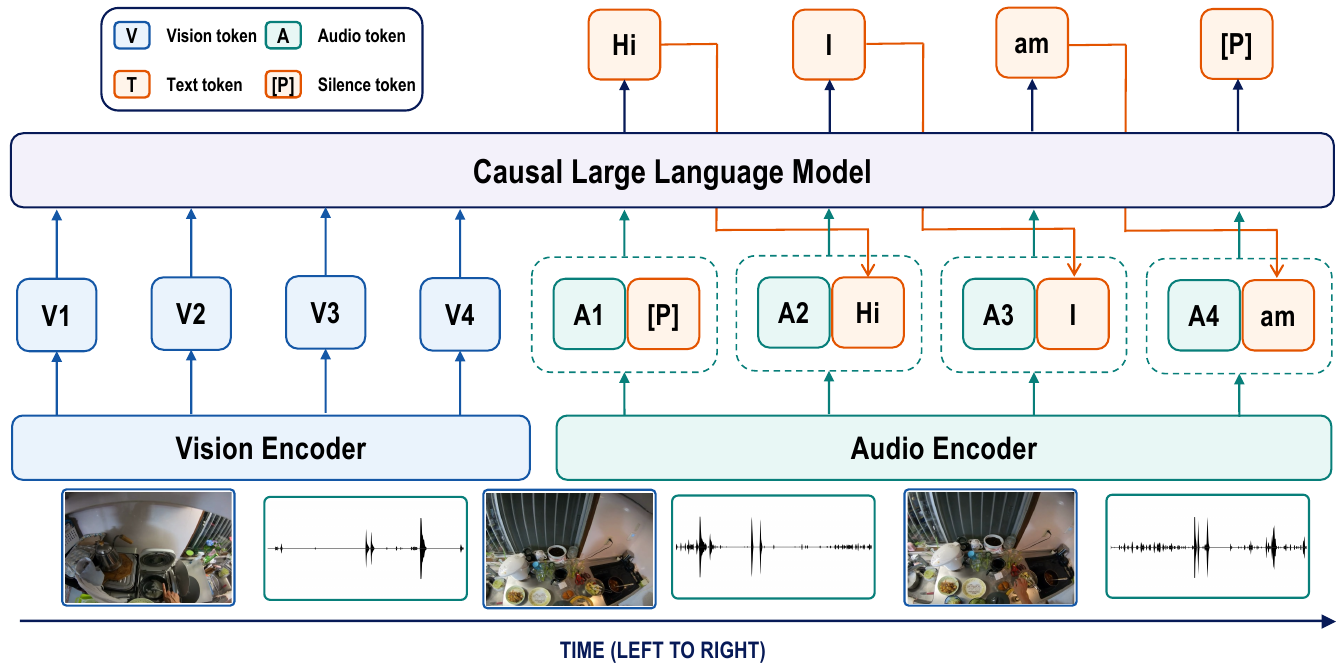}
    \caption{Aero Realtime architecture. Each audio state is fused with the preceding output embedding; $\mathtt{[P]}$ denotes silence.}
    \label{fig:model-architecture}
\end{figure*}
Together, these designs make Aero Realtime trainable and deployable as a long-running realtime system. The training stack exceeds 1,600 tokens per GPU per second on A100-40G GPUs. On four NVIDIA A6000 workstation GPUs, incremental inference maintains 84-ms median and 173-ms P95 processing lag over the first 20 minutes of a continuous video stream, remaining within 200~ms of the source timeline. On OVOBench, the compact 4B model achieves video-understanding performance comparable to several prior 7B methods~\citep{Shen2024LongVU,Wang2024Qwen2VL,Yao2025Timechat,Qian2025Dispider}, although it still trails the strongest baselines (Table~\ref{tab:ovobench-comparison}). To summarize, our contributions are threefold
\begin{itemize}
    \item We propose Aero Realtime, a 4B duplex multimodal architecture that aligns continuous perception, silence, and lexical generation on one temporal grid and jointly optimizes when and what to generate.
    \item We develop unified aligned-stream supervision for realtime and conventional video data, together with silence-aware optimization and modality-aware three-level parallel training.
    \item We introduce cache-valid delta inference for continuous input updates and validate sustained realtime execution over a 20-minute stream while retaining video-understanding capability comparable to several prior 7B methods.
\end{itemize}

\section{Related Work}
\label{sec:related-work}

\paragraph{Streaming video understanding.}

Streaming video-language models incrementally process long-running inputs. LiveCC interleaves frames with timestamped speech transcripts~\citep{Chen2025Livecc}, while StreamingVLM maintains a bounded KV cache for effectively unbounded streams~\citep{Xu2025Streamingvlm}. Simple Stream applies a recent-frame window to an off-the-shelf VLM~\citep{shen2026simplebaselinestreamingvideo}; MOSS-Video-Preview uses cross-attention pathways for concurrent perception and generation~\citep{wang2026mossvideopreviewrealtimevideounderstanding}; and Mage-VL uses codec-derived representations for efficient streaming~\citep{yang2026magevlefficientcodecnativestreaming}. VideoChat3 spans general, long-form, and streaming video understanding~\citep{li2026videochat3fullyopenvideo}. AURA learns continuous observation and proactive response within one VideoLLM~\citep{lu2026auraalwaysonunderstandingrealtime}, while JoyAI-VL-Interaction combines a vision-first interaction model with deployable agent delegation~\citep{yao2026joyaivlinteractionrealtimevisionlanguageinteraction}. Most methods retain per-step response generation or distinct trigger mechanisms, whereas Aero Realtime focuses on one slot-aligned input-output stream.

\paragraph{Duplex multimodal models.}

Duplex modeling has been explored most directly for spoken dialogue. Moshi jointly models separate user and assistant audio streams with time-aligned assistant text~\citep{defossez2024moshispeechtextfoundationmodel}; MoshiVis adds gated cross-attention to a static image~\citep{royer2025visionspeechmodelsteachingspeech}. Voxtral Realtime applies a stream-synchronous audio-plus-previous-token recurrence to ASR~\citep{mistralai2026voxtralrealtime}. Our recurrence follows this formulation but incorporates timestamped visual states and trains the output stream for proactive responses rather than transcription. Thinking Machines Lab's Interaction Models process continuous audio, video, and text with concurrent output and asynchronous background reasoning~\citep{thinkingmachines2026interactionmodels}. Aero Realtime uses an 80-ms grid where each audio slot predicts exactly one lexical or silence token, with visual states inserted by timestamp.

\section{Method}
\label{sec:method}

\subsection{Preliminaries: Multimodal Causal Transformers}
\label{sec:preliminaries}

Let $\mathcal{V}$, $\mathcal{A}$, and $\mathcal{T}$ denote visual observations, audio, and text, respectively. A conventional multimodal language model maps each supported modality into the hidden dimension of a causal Transformer,
\begin{equation}
    \mathbf{V}=P_v\!\left(f_v(\mathcal{V})\right),\qquad
    \mathbf{A}=P_a\!\left(f_a(\mathcal{A})\right),\qquad
    \mathbf{T}=E(\mathcal{T}),
    \label{eq:standard-multimodal-encoding}
\end{equation}
where $f_v$ and $f_a$ are modality encoders, $P_v$ and $P_a$ are projectors, and $E$ is the token embedding table. Existing multimodal LLMs serialize these states with text into a causal sequence~\citep{qwen3vl2025,an2026llavaonevision2nextgenerationperceptualintelligence}. In the turn-formatted setting, an input prefix $\mathbf{M}=\operatorname{Serialize}(\mathbf{V},\mathbf{A},\mathbf{T})$ precedes response tokens $t_{1:L}$, and training minimizes
\begin{equation}
    \mathcal{L}_{\mathrm{turn}}
    =-\sum_{i=1}^{L}\log p_\theta
    \!\left(t_i\mid \mathbf{M},t_{<i}\right).
    \label{eq:turn-objective}
\end{equation}
Unlike this input-prefix/output-suffix formulation, Aero Realtime advances perception and generation on a shared timeline.

\begin{figure*}[t]
    \centering
    \begin{minipage}[c]{0.68\textwidth}
        \centering
        \includegraphics[width=\linewidth]{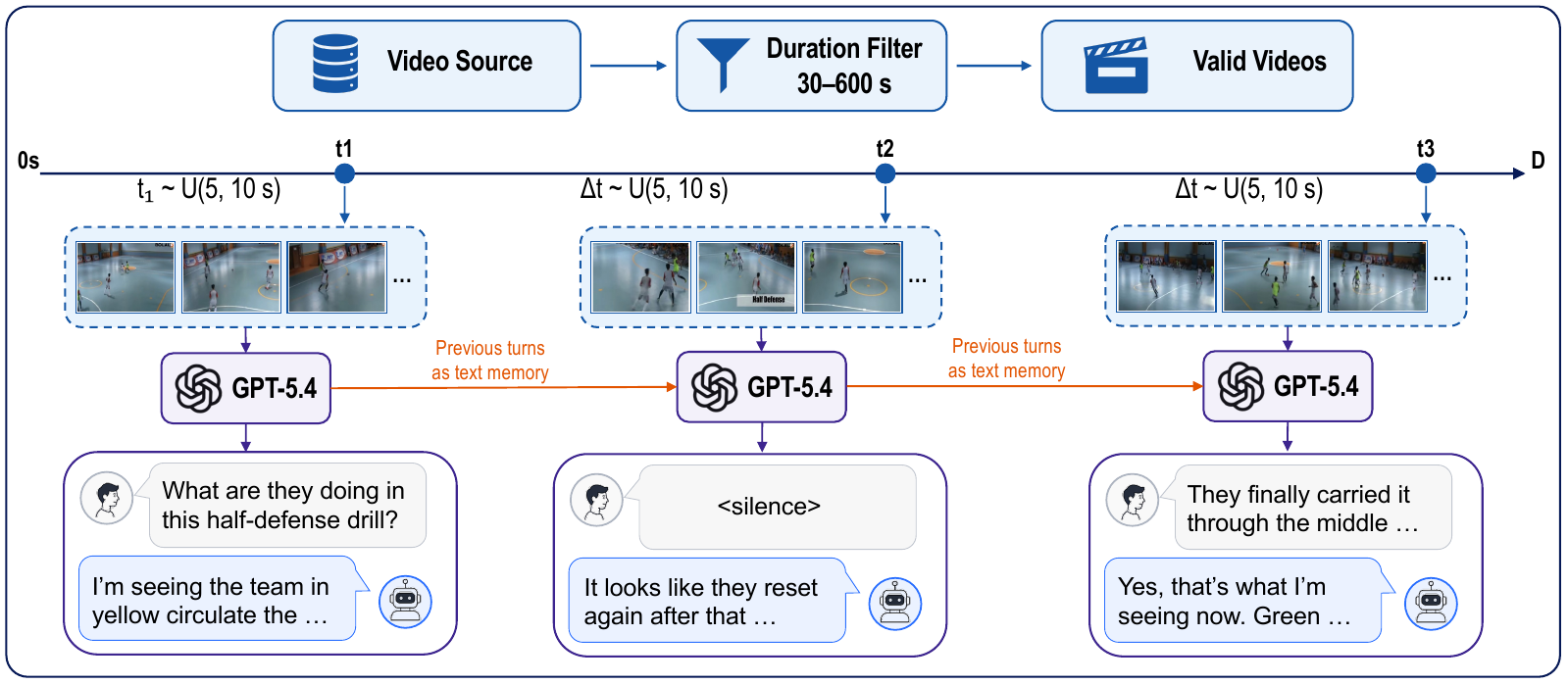}
    \end{minipage}\hfill
    \begin{minipage}[c]{0.3\textwidth}
        \centering
        \includegraphics[width=\linewidth]{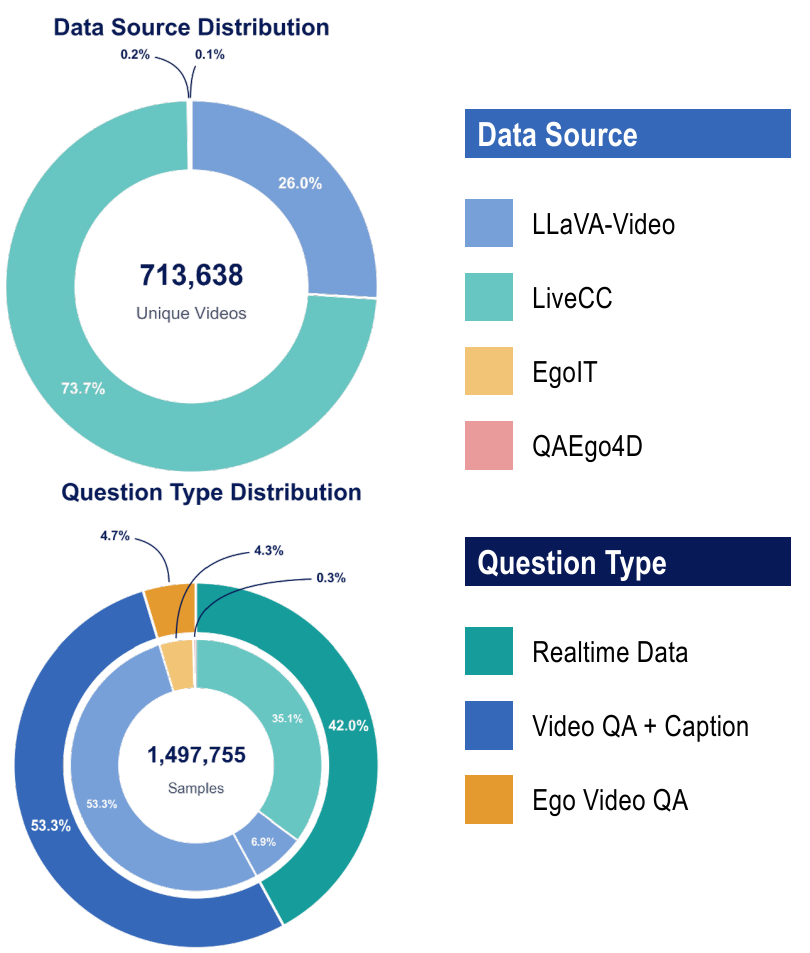}
    \end{minipage}
    \caption{Realtime QA construction and training-data composition. Generation uses only past visual context and turns; source shares use unique videos and question-type shares use deduplicated samples.}
    \label{fig:data-pipeline-distribution}
\end{figure*}
\subsection{Model Architecture}
\label{sec:model-architecture}

Figure~\ref{fig:model-architecture} presents the overall architecture of Aero Realtime. We preserve modality-specific vision and audio towers but replace turn-level conditioning with a dense text stream aligned to the audio timeline. At each audio slot, the current audio representation is combined with the preceding text-stream token embedding before entering the causal language model, which predicts the next aligned lexical token or the silence token $\mathtt{[P]}$.

Aero Realtime divides continuous time into $N$ audio slots. The audio encoder and projector temporally downsample the waveform such that each projected audio hidden state represents approximately 80~ms of input. Let $a_t$ denote the audio segment at slot $t$, $v_j$ the $j$-th sampled video frame with timestamp $\tau_j$, and $y_t$ the output token aligned with slot $t$. The modality-specific representations are
\begin{equation}
    z_j^v=P_v\!\left(f_v(v_j)\right),
    \qquad
    z_t^a=P_a\!\left(f_a(a_t)\right),
    \label{eq:modality-encoding}
\end{equation}
where $z_j^v,z_t^a\in\mathbb{R}^d$ match the hidden dimension of the language model. We denote the visual context available at audio slot $t$ as
\begin{equation}
    \mathcal{Z}_{\leq t}^v
    =\{z_j^v\mid \tau_j\leq \tau_t\},
\end{equation}
where $\tau_t$ is the wall-clock timestamp of the audio slot.

Each audio slot has exactly one output token,
\begin{equation}
    y_t\in\mathcal{V}_{\mathrm{text}}\cup\{\mathtt{[P]}\},
\end{equation}
where $\mathtt{[P]}$ is a special silence token with a learned embedding $E(\mathtt{[P]})\in\mathbb{R}^d$. A lexical token indicates that the response advances at the current slot, whereas $\mathtt{[P]}$ indicates that the model remains silent. Our causal audio-text recurrence follows the realtime formulation of Voxtral Realtime~\citep{mistralai2026voxtralrealtime}; Aero Realtime extends it to multimodal generation by incorporating timestamp-aligned visual states into the same causal stream. To carry the output state along the audio timeline, we fuse the current audio hidden state with the preceding output-token embedding,
\begin{equation}
    e_t^a=z_t^a+E(y_{t-1}),
    \qquad y_0=\mathtt{[P]}.
    \label{eq:audio-text-fusion}
\end{equation}
Thus, listening context and response history are represented at every slot. An $L$-token response occupies $L$ consecutive 80-ms slots while perception continues, yielding a maximum lexical rate of 12.5 tokens/s without accumulating an input backlog.

Visual states remain independent tokens, whereas each audio state shares a position with the preceding text state. Following Qwen2.5-Omni and Qwen3-Omni~\citep{xu2025qwen25omnitechnicalreport,xu2025qwen3omnitechnicalreport}, timestamped visual tokens are interleaved with the fused audio-text slots.

Let $h_t$ denote the Transformer hidden state associated with the fused token $e_t^a$. It is conditioned on all preceding fused audio-text tokens and all visual tokens available by time $t$,
\begin{equation}
    h_t=\operatorname{Transformer}
    \!\left(\mathcal{Z}_{\leq t}^v,e_{\leq t}^a\right).
    \label{eq:aligned-transformer}
\end{equation}
The model predicts the aligned output using the original language-model head,
\begin{equation}
    p_\theta(y_t\mid \mathcal{Z}_{\leq t}^v,a_{\leq t},y_{<t})
    =\operatorname{softmax}(W_{\mathrm{lm}}h_t),
    \label{eq:aligned-generation}
\end{equation}
Predicting either a lexical token or $\mathtt{[P]}$ jointly models \emph{what} to generate and \emph{when}, without a separate response gate.

\subsubsection{Training Objective}

The architecture retains standard next-token cross entropy, but changes its conditioning context from the turn-level prefix in Equation~\ref{eq:turn-objective} to the aligned multimodal history,
\begin{equation}
    \mathcal{L}_{\mathrm{stream}}
    = -\sum_{t=1}^{N} m_t
      \log p_\theta\!\left(y_t\mid x_{\leq t},y_{<t}\right),
    \label{eq:stream-loss}
\end{equation}
where $x_{\leq t}=(\mathcal{Z}_{\leq t}^v,a_{\leq t})$ denotes observations causally available through slot $t$, and $m_t$ masks excluded positions. Packed samples are shifted independently to prevent loss across sequence boundaries. Lexical tokens and $\mathtt{[P]}$ share the LM head, requiring no auxiliary timing loss.

\begin{figure}[t]
    \centering
    \includegraphics[width=0.48\textwidth]{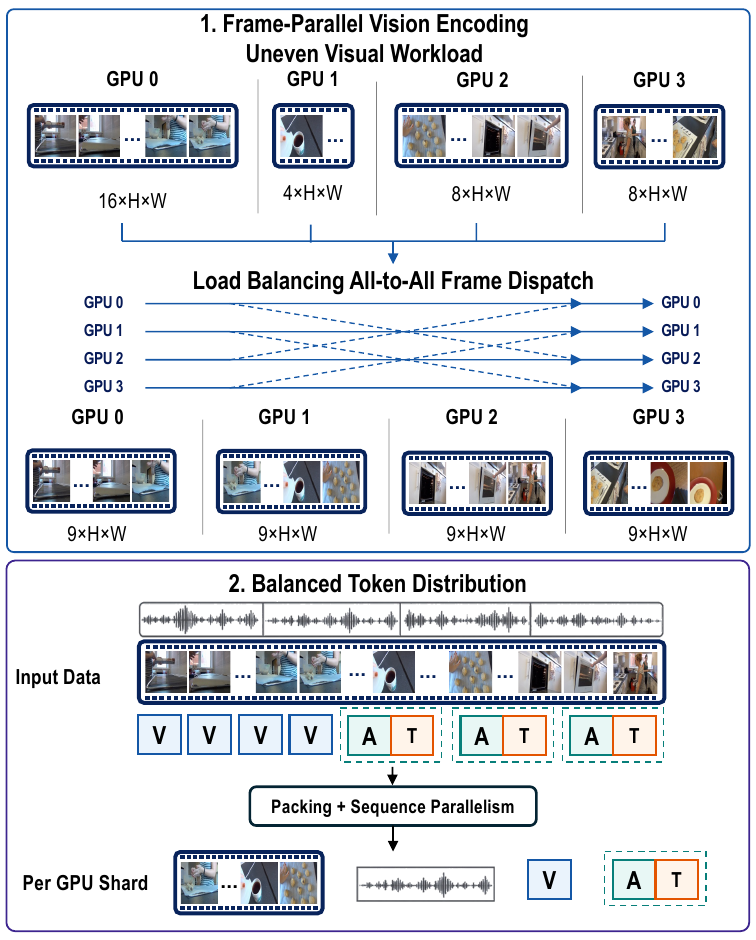}
    \caption{Three-level training parallelism: workload-balanced frame parallelism for vision and Ulysses sequence parallelism for audio and language.}
    \label{fig:training-infrastructure}
\end{figure}
\subsection{Data Curation}
\label{sec:data-curation}

Figure~\ref{fig:data-pipeline-distribution} summarizes our causal realtime QA generation pipeline and the composition of the resulting training mixture.

\subsubsection{Data Construction Pipeline}

We first retain videos between 30 and 600 seconds. For each video, we sample an initial timestamp uniformly between 5 and 10 seconds and advance through the remaining video with intervals drawn from the same range. At each timestamp, GPT-5.4 receives only the video prefix available at that time together with all previously generated QA turns as text memory, and generates a new temporally grounded QA pair. Repeating this causal procedure produces a time-ordered sequence of realtime interactions without exposing future frames. This pipeline yields 103k realtime QA samples.

Our training mixture also includes existing video corpora converted into the aligned stream format: video QA and caption data from LLaVA-Video~\citep{Zhang2024LLaVAVideo} and LiveCC~\citep{Chen2025Livecc}, and egocentric video QA from EgoIT~\citep{yang2026egolifeegocentriclifeassistant} and QAEgo4D~\citep{patel2025advancingegocentricvideoquestion}. Sources carrying temporal annotations are converted as realtime data, and the remaining question-answer and caption samples are converted as conventional instruction data.

We preserve source identifiers and split metadata during conversion and remove duplicate source paths when assembling the mixture. We do not claim that this path-level procedure establishes content-level independence from every evaluation video. In particular, because the training mixture and OVOBench draw on related egocentric-video sources, possible source overlap remains a limitation discussed in Section~\ref{sec:conclusion}.

\subsubsection{Aligned-Stream Conversion}

For temporally annotated examples, we map each response to the first audio slot at or after its timestamp, place the tokenized response in the following slots, and supervise all unoccupied positions as $\mathtt{[P]}$. Conventional video QA requires only a simple adaptation: we keep the question as context, append silent audio slots after the video, and place the answer tokens at the corresponding output positions. Samples without video retain the standard conversation-template supervision.

\subsection{Delta-Based Realtime Inference}
\label{sec:realtime-inference}

Using vLLM-Omni resumable requests~\citep{yin2026vllmomni}, each 80-ms update appends one audio feature and any timestamped visual or structural tokens, then emits one token $y_t$. The sampled token is fused into the next audio update rather than retained as an ordinary generated suffix:
\begin{equation}
    e_{t+1}=P_a\!\left(f_a(a_{t+1})\right)+E(y_t).
\end{equation}

The scheduler removes the sampled suffix, appends only the new multimodal delta, and reuses the KV-cached prefix, so only new positions are evaluated. For audio-only continuation,
\begin{equation}
    (y_t,\mathrm{KV}_t)
    =F_\theta\!\left(x_t,y_{t-1},\mathrm{KV}_{t-1}\right).
\end{equation}
\begin{figure}[t]
    \centering
    \includegraphics[width=0.45\textwidth]{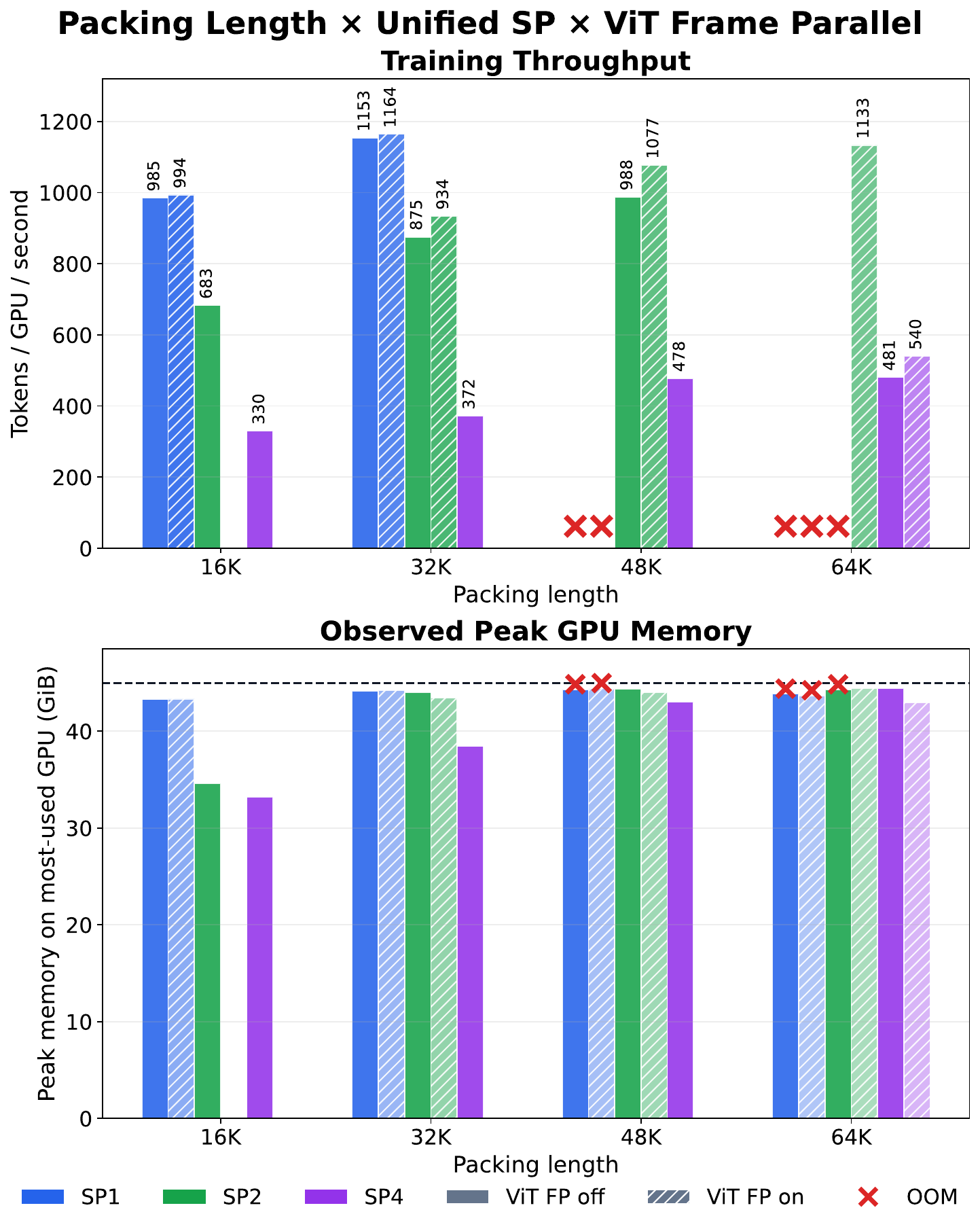}
    \caption{Throughput and peak memory across packing lengths and parallel settings on four A6000 GPUs. Hatching enables ViT frame parallelism; crosses denote OOM.}
    \label{fig:parallel-ablation}
\end{figure}
\subsection{Three-Level Parallel Training Infrastructure}
\label{sec:training-infrastructure}

Figure~\ref{fig:training-infrastructure} shows our modality-aware three-level strategy. \textbf{Workload-balanced frame parallelism.} Text-token packing alone does not balance the vision tower: packed samples contain videos with different numbers and resolutions of frames, while conventional LM-only sequence parallelism replicates the same visual encoding on every rank. For frame $f$, we estimate its visual workload as $w_f=H_fW_f$ from the patch grid and assign frames to $S$ sequence-parallel ranks by approximately minimizing
\begin{equation}
    \max_{r\in\{1,\ldots,S\}}\sum_{f:\,\pi(f)=r} w_f,
    \label{eq:frame-load-balance}
\end{equation}
where $\pi(f)$ is the rank processing frame $f$. A deterministic locality-aware redistribution moves frames from overloaded to underloaded ranks before the vision forward pass and gathers their features afterward. This frame-level partition equalizes vision compute, avoids replicated visual activations, and reduces both stragglers and peak memory.

\textbf{Unified packed-sequence parallelism.} After on-the-fly packing, we remove padding and partition the audio sequence and the final multimodal language sequence across the same SP group. Both towers use DeepSpeed Ulysses~\citep{jacobs2023deepspeedulyssesoptimizationsenabling}, which exchanges sequence and attention-head dimensions through all-to-all communication while preserving sample boundaries. Combining workload-balanced frame parallelism with audio- and LM-sequence parallelism distributes all three dominant components; applying SP only to the LM would leave each rank repeatedly encoding the same frames and audio, limiting memory savings for long multimodal streams.

\begin{table*}[t]
\centering
\small
\setlength{\tabcolsep}{4.0pt}
\begin{tabular}{lcccccccc}
\toprule
Model & LM & Input Mode & Duplex I/O & Native Proactive & Realtime & Backward & Forward & Avg. \\
\midrule
Qwen2.5-VL~\citep{bai2025qwen25vltechnicalreport} & 7B & Offline & \xmark & \xmark & 59.90 & 44.70 & -- & -- \\
LLaVA-OneVision~\citep{li2024llavaonevisioneasyvisualtask} & 7B & Offline & \xmark & \xmark & 64.00 & 43.70 & 50.50 & 52.73 \\
InternVL2~\citep{Chen2024InternVL2} & 8B & Offline & \xmark & \xmark & 60.40 & 43.40 & 46.60 & 50.13 \\
LLaVA-Video~\citep{Zhang2024LLaVAVideo} & 7B & Offline & \xmark & \xmark & 63.50 & 40.40 & \textbf{54.82} & 52.91 \\
Qwen2-VL~\citep{Wang2024Qwen2VL} & 7B & Offline & \xmark & \xmark & 60.70 & 48.60 & 48.74 & 52.68 \\
LongVU~\citep{Shen2024LongVU} & 7B & Offline & \xmark & \xmark & 57.40 & 39.50 & 47.50 & 48.13 \\
\midrule
VideoLLM-online~\citep{Chen2024VideoLLMOnline} & 8B & Online & \xmark & \xmark & 20.80 & 17.70 & -- & -- \\
Flash-VStream~\citep{Zhang2025Flash} & 7B & Online & \xmark & \xmark & 28.40 & 27.40 & 45.09 & 33.63 \\
Dispider~\citep{Qian2025Dispider} & 7B & Online & \xmark & \xmark & 54.60 & 36.10 & 34.72 & 41.81 \\
TimeChat-Online~\citep{Yao2025Timechat} & 7B & Online & \xmark & \xmark & 61.90 & 41.70 & 36.70 & 46.77 \\
StreamForest~\citep{Zeng2025Streamforest} & 7B & Online & \xmark & \xmark & 61.20 & \textbf{52.00} & 53.49 & 55.56 \\
Streamo~\citep{Xia2025StreamingVideo} & 7B & Online & \xmark & \cmark & 66.00 & 46.10 & 54.77 & \textbf{55.62} \\
HERMES~\citep{Zhang2026Hermes} & 7B & Online & \xmark & \xmark & \textbf{69.00} & 49.40 & -- & -- \\
\midrule
\textbf{Aero Realtime} & \textbf{4B} & \textbf{Online} & \cmark & \cmark & 61.49 & 44.07 & 39.36 & 48.31 \\
\bottomrule
\end{tabular}
\caption{OVOBench track averages and interaction properties. Duplex I/O admits observations during generation; Native Proactive models silence and text in one autoregressive objective. Properties follow public sources; column maxima are bold.}
\label{tab:ovobench-comparison}
\end{table*}
\section{Experiments}
\label{sec:experiments}

\subsection{Experimental Setup}
\label{sec:experimental-setup}

\paragraph{Model initialization.}
Aero Realtime is initialized from Qwen3-VL-4B-Instruct~\citep{bai2025qwen3vltechnicalreport}, whose vision tower and language model we retain. The audio tower is initialized from Qwen3-Omni~\citep{xu2025qwen3omnitechnicalreport} and adopts the temporal downsampling scheme of Qwen2-Audio~\citep{chu2024qwen2audiotechnicalreport}, so that each projected audio token corresponds to an 80-ms chunk and matches the audio slot grid of Section~\ref{sec:model-architecture}.

\paragraph{Training configuration.}
We train with LMMs-Engine~\citep{lmms_engine2025} using sequence-parallel size 4 and data-parallel size 8, for a world size of 32. Optimization uses fused AdamW, bf16 precision, a constant $5\!\times\!10^{-5}$ learning rate with 5\% warmup, and 65k-token packed sequences. Padding removal and Liger kernels~\citep{hsu2025ligerkernel} improve efficiency; because packing is online, training length is measured in consumed tokens rather than epochs.

\paragraph{Training stages.}
Training proceeds in two stages. The first stage uses the full training mixture to adapt the model to the aligned input-output stream distribution and consumes 2B tokens. The second stage strengthens response quality and video understanding, using our generated realtime data together with a subset of LLaVA-Video and QAEgo4D, and consumes 1B tokens.

\paragraph{Evaluation.}
All benchmark results use LMMs-Eval~\citep{zhang2024lmmsevalrealitycheckevaluation,lmms_eval2024}. Training uses seed 42, and each reported configuration is evaluated from one training run. Complete optimization, data-sampling, parallelism, and hardware settings are provided in the supplementary appendix.

\subsection{Video Understanding Performance}
\label{sec:ovobench-performance}

We evaluate streaming video understanding on OVOBench~\citep{Li2025Ovo}. Table~\ref{tab:ovobench-comparison} reports the average score for each track together with the interaction properties of each architecture. Aero Realtime uses a 4B language model and achieves 61.49 on Realtime, 44.07 on Backward, and 39.36 on Forward. It does not match the strongest baselines: it trails the best online models on all three tracks, and its Forward score is the lowest among the compared models. However, it is the only evaluated architecture that admits new observations into an active decoding sequence, while also modeling silence and lexical generation within one autoregressive output stream.

We hypothesize that this gap reflects several factors: the distribution shift from turn-formatted pretraining to dense aligned streams, the limited amount of native realtime supervision, and the large proportion of silence-aligned positions. The current experiments do not isolate their individual contributions. The results therefore demonstrate retained video-understanding capability under the aligned formulation rather than state-of-the-art benchmark accuracy.

\subsection{Realtime Latency and Response Time}
\label{sec:realtime-latency}

We evaluate whether Aero Realtime can keep its input and output streams fully overlapped under continuous audio-video input. Unlike a turn-based system, generation never closes an input turn: each 80-ms slot admits the next audio chunk, an optional video frame, and the previous slot's output token into the same active autoregressive sequence. The model therefore continues ingesting new observations while it is deciding whether to remain silent or emit lexical output; speaking does not pause perception or start a separate generation request.

\begin{figure}[t]
\centering
\includegraphics[width=\columnwidth]{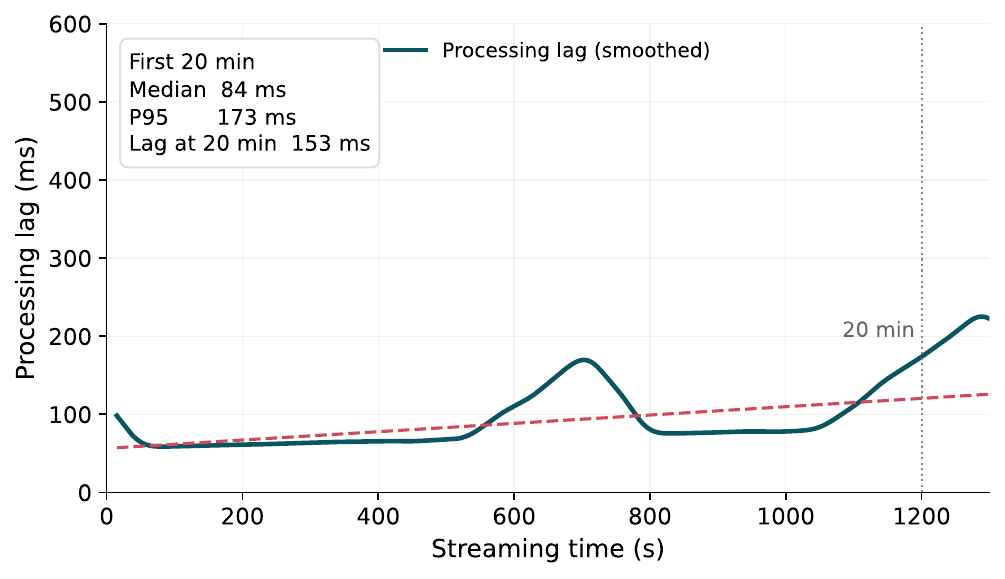}
\caption{Long-horizon processing lag under continuous audio-video streaming.}
\label{fig:always-on-time}
\end{figure}

Figure~\ref{fig:always-on-time} reports the wall-clock completion lag of each processed interval relative to its source timestamp for a 30-minute Video-MME clip, streamed at its native wall-clock rate on four NVIDIA A6000-45G GPUs. The curve uses a 120-second Gaussian smoothing window to expose the long-horizon trend. For the first 20 minutes, Aero Realtime maintains a median lag of 84\,ms, P95 lag of 173\,ms, and 153\,ms lag at the 20-minute boundary. Thus, even as the active sequence grows continuously, fully overlapping input and output keeps the system within 200\,ms of the source stream for 20 minutes. This demonstrates sustained realtime processing rather than short-clip throughput or offline catch-up.

\begin{table}[t]
\centering
\small
\setlength{\tabcolsep}{3.4pt}
\begin{tabular}{lrrrrrr}
\toprule
Streaming time & 30\,s & 1\,min & 5\,min & 10\,min & 15\,min & 20\,min \\
\midrule
Processing lag (ms) & 37 & 37 & 39 & 95 & 56 & 153 \\
\midrule
\multicolumn{5}{l}{Median / P95 over first 20\,min} & \multicolumn{2}{r}{84 / 173\,ms} \\
\bottomrule
\end{tabular}
\caption{Processing lag at selected streaming checkpoints.}
\label{tab:streaming-lag-checkpoints}
\end{table}

\subsection{Effect of Silence-Label Masking}
\label{sec:silence-mask-ablation}

Silence labels dominate the aligned stream and can overwhelm sparse lexical targets. We define $r$ as the fraction of $\mathtt{[P]}$ targets independently excluded from the loss; lexical targets are never masked. Without masking, the model remains almost entirely silent, yielding only 6.40 average on OVOBench. Although $r=0.70$ recovers Forward performance, $r=0.95$ produces more balanced results, raising the average to 42.36 while retaining 5\% of silence labels for Stage~1.

\begin{table}[t]
\centering
\small
\setlength{\tabcolsep}{3.5pt}
\resizebox{0.45\textwidth}{!}{%
\begin{tabular}{lcccc}
\toprule
& \multicolumn{4}{c}{OVOBench} \\
\cmidrule(lr){2-5}
Setting & Realtime & Backward & Forward & Avg. \\
\midrule
No masking ($r=0.00$) & 0.36 & 0.96 & 17.87 & 6.40 \\
\rowcolor{oursrow} $+$ Masking ($r=0.70$) & 9.03 & 12.01 & \textbf{47.90} & 22.98 \\
\rowcolor{oursrow} $+$ Masking ($r=0.95$, Stage 1) & 58.03 & 33.94 & 35.11 & 42.36 \\
\rowcolor{oursrow} $+$ Stage 2 & \textbf{61.49} & \textbf{44.07} & 39.36 & \textbf{48.31} \\
\bottomrule
\end{tabular}
}
\caption{Effects of silence-label masking and staged training on OVOBench.}
\label{tab:training-ablation}
\end{table}

\subsection{Training-Stage Ablations}
\label{sec:stage-ablation}

The two stages separate interface adaptation from capability refinement. Stage~1 uses the full data mixture to teach the pretrained turn-based model the new slot-aligned interface, including silence prediction and continuous lexical output. Once this behavior is established, Stage~2 uses our realtime data with selected LLaVA-Video and QAEgo4D examples to reinforce response quality and temporal understanding without relearning the interface from scratch. It improves Realtime from 58.03 to 61.49 and Backward from 33.94 to 44.07, raising the overall average by 5.95 points to 48.31. The larger Backward gain suggests that the focused second-stage mixture particularly improves reasoning over previously observed events, while preserving the realtime behavior learned in Stage~1.

\subsection{Training Infrastructure Ablation}
\label{sec:infrastructure-ablation}

Figure~\ref{fig:parallel-ablation} studies packing length, unified sequence-parallel degree, and ViT frame parallelism on four A6000 GPUs. Longer packing improves throughput only when sufficient sequence parallelism keeps activation memory tractable: low SP degrees run out of memory, whereas increasing SP reduces peak memory with a modest communication overhead. ViT frame parallelism consistently recovers throughput by balancing visual work and avoiding replicated frame encoding. Guided by this trade-off, our full training uses an $\mathrm{SP}=4\times\mathrm{DP}=8$ topology over 32 ranks. The system sustains over 1,600 tokens per GPU per second with limited parallelism overhead, allowing both training stages to finish within one day. These results show that the three-level design scales long packed multimodal streams beyond the four-GPU ablation setting.



\section{Conclusion}
\label{sec:conclusion}

We presented Aero Realtime, which aligns streaming observations, silence, and lexical output on one temporal grid and preserves KV-cache reuse across continuous updates. Our data, training, and serving recipe demonstrates feasible duplex, natively proactive, and hardware-aligned interaction. Aero Realtime remains an architectural exploration: its 80-ms grid caps output at 12.5 tokens per second, and it trails the strongest video LLMs on OVOBench. Moreover, latency reflects deployment policy rather than isolated forward speed, while path-level deduplication does not guarantee content-level independence from every evaluation video. Adaptive-rate decoding, more native realtime supervision, and stronger contamination audits remain future work.

\section*{Acknowledgments}
This work was supported by the Hong Kong Research Grants Council General Research Fund (Grant Nos.~17202422, 17212923, and 17215025), Theme-based Research Scheme (Grant No.~T45-701/22-R), and Strategic Topics Grant (Grant No.~STG3/E-605/25-N). Part of this research was conducted in the JC STEM Lab of Robotics for Soft Materials, funded by The Hong Kong Jockey Club Charities Trust.

\bibliography{aaai2027}

\clearpage
\appendix

\maketitle

\section{Training Configuration}

Table~\ref{tab:appendix-hyperparameters} lists the final hyperparameters used for both training stages. The two stages share all optimization and parallelism settings and differ only in the number of training steps and in the data mixture, as described in the main paper.

\begin{table}[h]
\centering
\small
\begin{tabular}{ll}
\toprule
Hyperparameter & Value \\
\midrule
Optimizer & AdamW (fused) \\
Learning rate & $5\times10^{-5}$ \\
Learning-rate schedule & Constant \\
Warmup ratio & 0.05 \\
Precision & bf16 (tf32 matmul) \\
Gradient checkpointing & Enabled \\
Per-device batch size & 1 \\
Gradient accumulation & 1 \\
Stage 1 steps & 4000 \\
Stage 2 steps & 2000 \\
\midrule
Packing length & 65536 tokens \\
Packing strategy & Balanced (10 buckets) \\
Padding removal & Enabled \\
Sharding & FSDP2 \\
Sequence-parallel degree & 4 \\
Attention implementation & FlashAttention-2 \\
\midrule
Frame rate & 1 fps \\
Max frames per video & 64 \\
Max visual pixels & 360448 \\
Min visual pixels & 28800 \\
Silence-label masking ratio & 0.95 \\
\bottomrule
\end{tabular}
\caption{Final hyperparameters for both training stages.}
\label{tab:appendix-hyperparameters}
\end{table}

We did not perform a systematic hyperparameter search. The learning rate and the learning-rate schedule were the only settings varied during development, and the reported configuration was selected from a small number of preliminary runs. All remaining values were fixed a priori.

\section{Training-Stage Data Mixtures}

Training length is defined by consumed tokens because samples are packed on the fly. Stage~1 uses the complete mixture to adapt the turn-based initialization to aligned-stream generation. Stage~2 starts from the Stage~1 checkpoint and uses a focused mixture to strengthen response quality and video understanding.

\begin{table}[h]
\centering
\small
\begin{tabular}{p{0.12\linewidth}p{0.62\linewidth}p{0.14\linewidth}}
\toprule
Stage & Data mixture & Token budget \\
\midrule
Stage 1 & Generated realtime QA; LLaVA-Video QA/caption data; LiveCC; EgoIT; QAEgo4D & 2B \\
Stage 2 & Generated realtime QA; selected LLaVA-Video data; QAEgo4D & 1B \\
\bottomrule
\end{tabular}
\caption{Data mixtures and token budgets for the two training stages.}
\label{tab:appendix-stage-data}
\end{table}

\section{Computing Infrastructure}

All training runs use 4 nodes with 8 NVIDIA A100-40G GPUs each, for a total of 32 GPUs. The software stack is CUDA 13.2 and PyTorch 2.11, with LMMs-Engine for training and LMMs-Eval for evaluation. The parallel-training ablation and long-horizon latency experiment use 4 NVIDIA A6000-45G GPUs. CPU model, host memory, and operating system version are not recorded.

\section{Random Seeds}

All training runs use a fixed random seed of 42, which controls data shuffling, packing order, and model initialization of newly added parameters. We report results from a single run per configuration and therefore do not report variance across seeds.

\section{Qualitative Proactive Examples}

The following examples are taken directly from the saved realtime runs. The image shows the current stream state, while the text is emitted by Aero Realtime as the stream continues; no separate answer-generation turn is opened.

\section{Realtime Benchmark Data Samples}

The benchmark samples pair a continuous video with time-aligned user turns and assistant responses. Each row below shows the video state at the annotated interaction time and the corresponding supervision.
\begin{figure*}[p]
\centering
\begin{minipage}{0.92\textwidth}
\begin{AIbox}[width=\linewidth]{Proactive response at 00:32}
\begin{center}
\begin{tabular}{@{}c@{\hspace{12pt}}p{0.56\textwidth}@{}}
\includegraphics[width=0.30\textwidth]{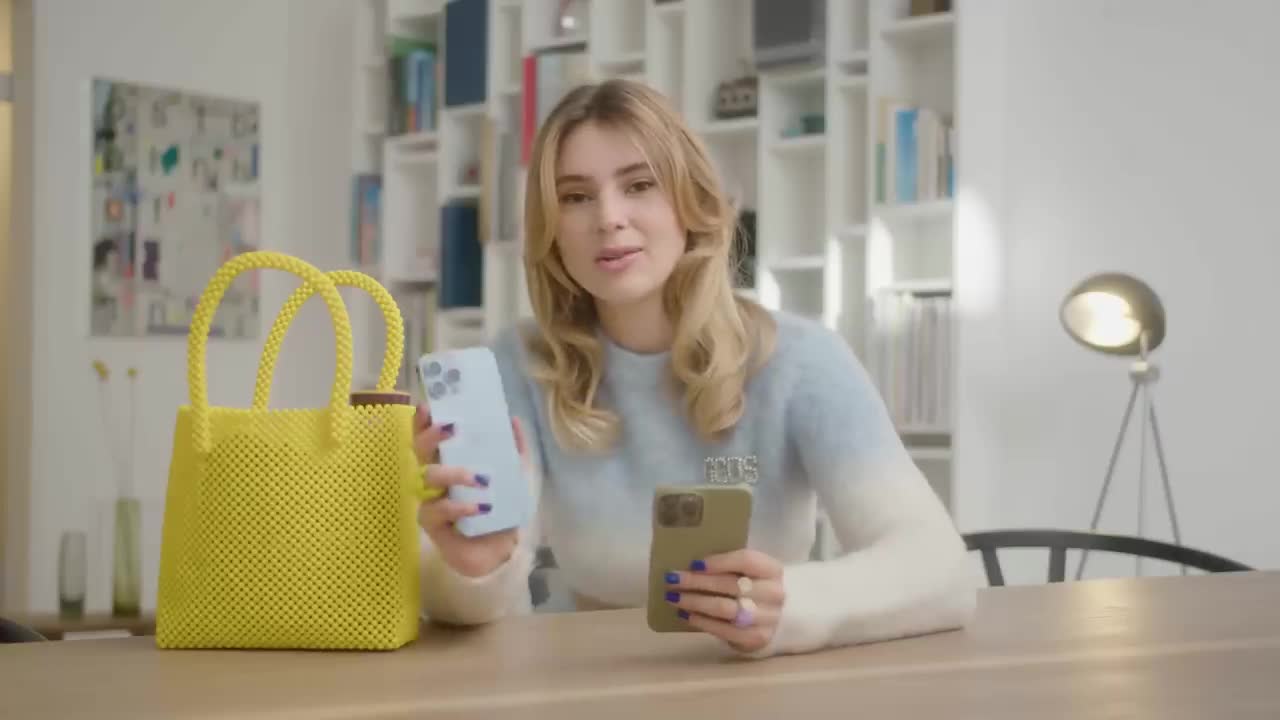} & \textbf{00:30} \quad I can see a person seated at a table with a bright yellow bag on the left side of the frame. The person is \tabularnewline[4pt]
\includegraphics[width=0.30\textwidth]{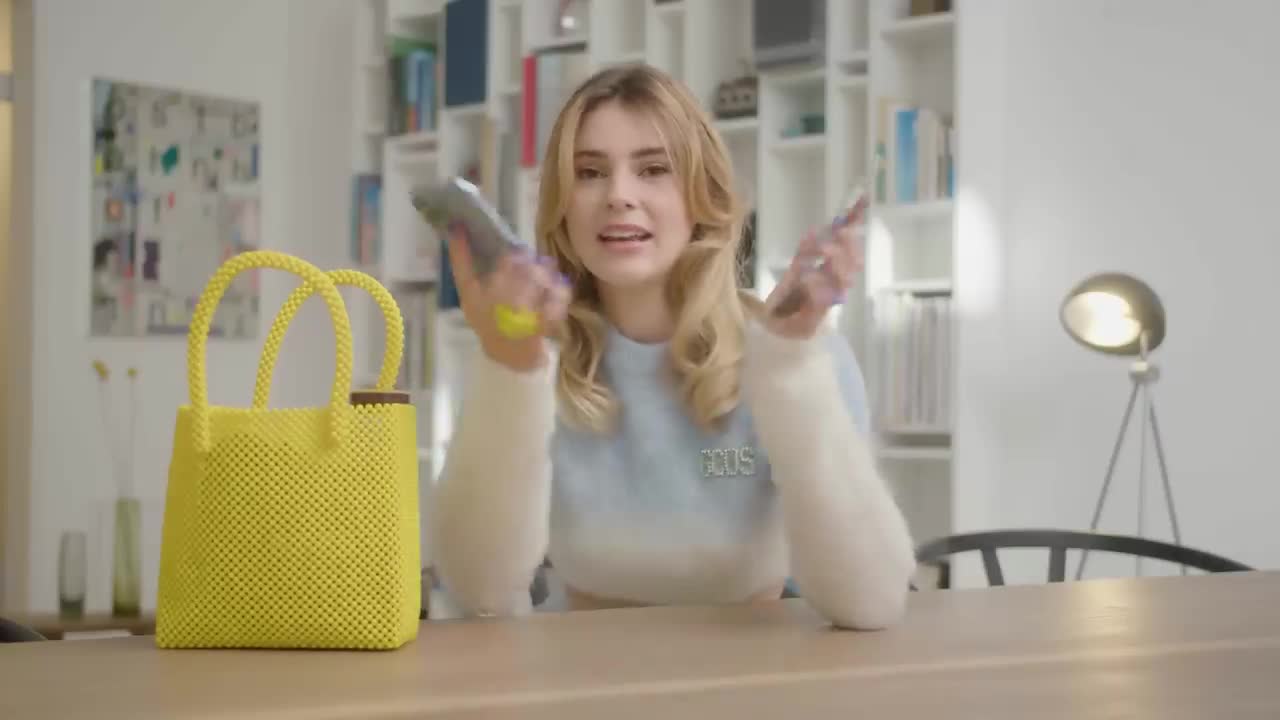} & \textbf{00:32} \quad gesturing with their hands and appears to be talking. The background includes a bookshelf and a floor lamp. The person is \tabularnewline[4pt]
\includegraphics[width=0.30\textwidth]{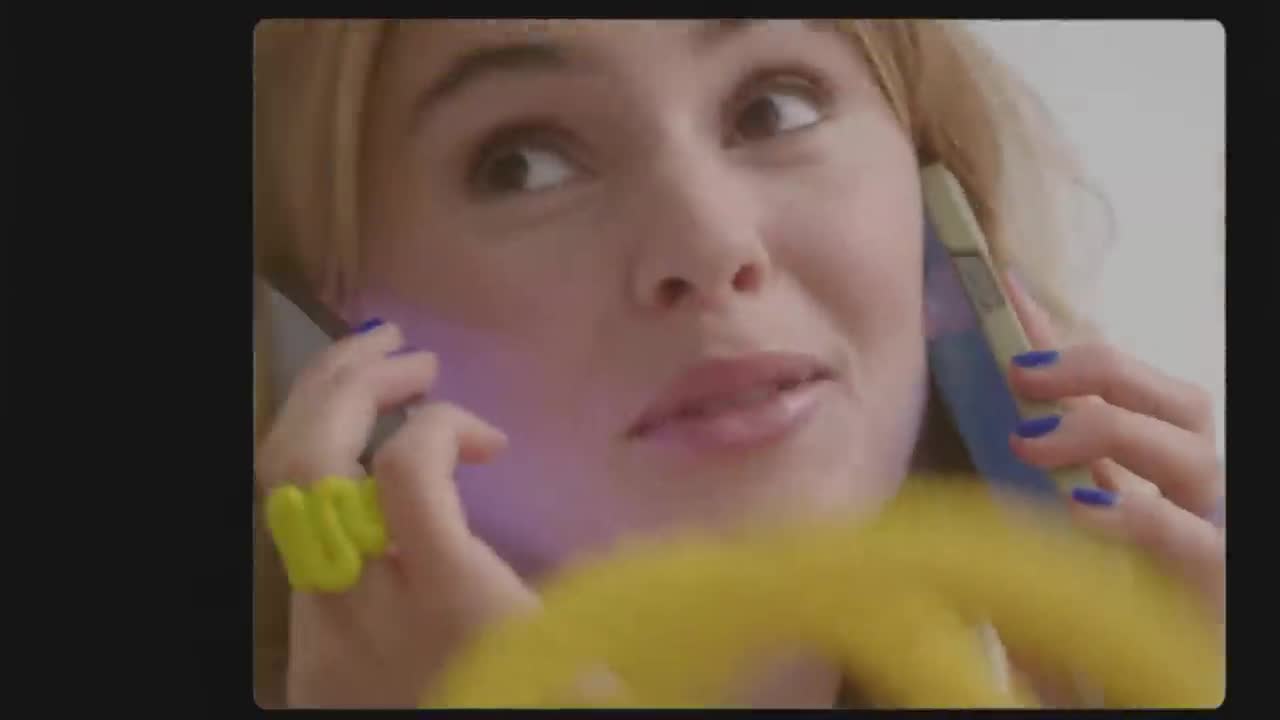} & \textbf{00:34} \quad still seated at the table, continuing to gesture and talk. The person is still seated at the table, continuing to gesture and \tabularnewline[4pt]
\includegraphics[width=0.30\textwidth]{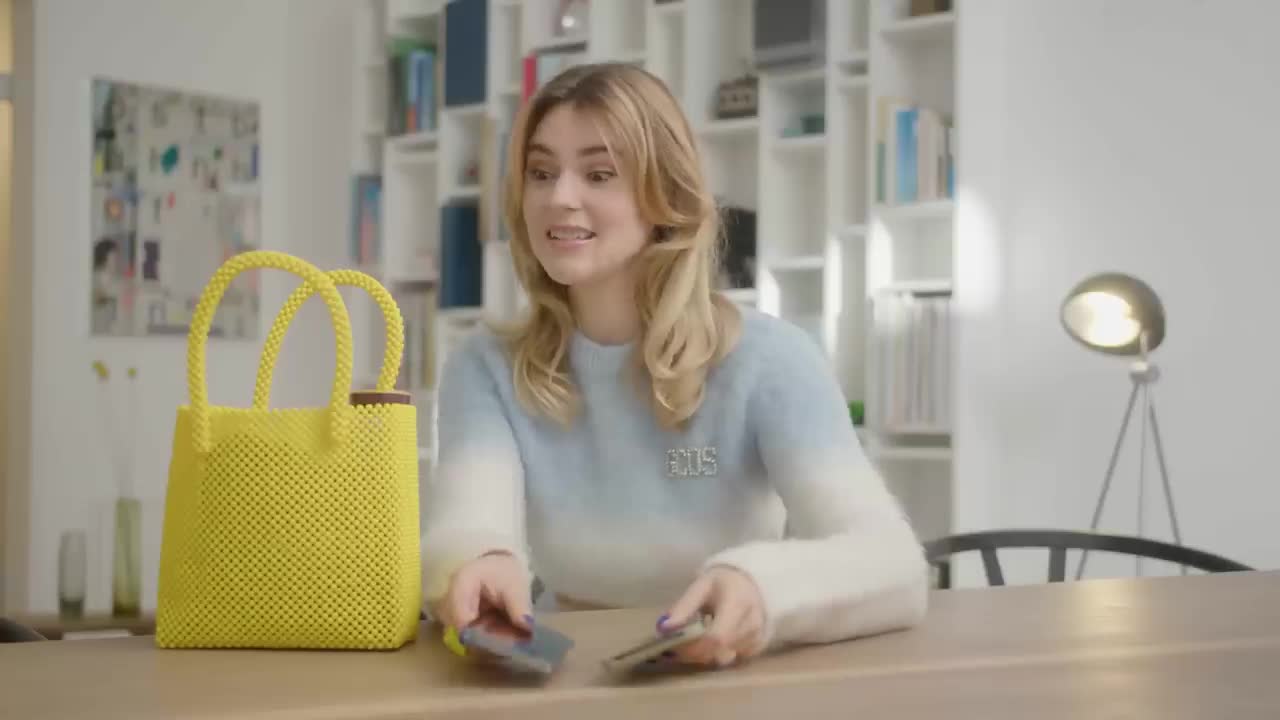} & \textbf{00:36} \quad talk. The bright yellow bag remains on the left side of the frame, with the bookshelf and floor lamp in the background. \\
\end{tabular}
\end{center}
\textbf{Standing instruction:} Describe what is happening now.
\end{AIbox}
\end{minipage}
\end{figure*}

\begin{figure*}[p]
\centering
\begin{minipage}{0.92\textwidth}
\begin{AIbox}[width=\linewidth]{Proactive response at 00:12}
\begin{center}
\begin{tabular}{@{}c@{\hspace{12pt}}p{0.56\textwidth}@{}}
\includegraphics[width=0.30\textwidth]{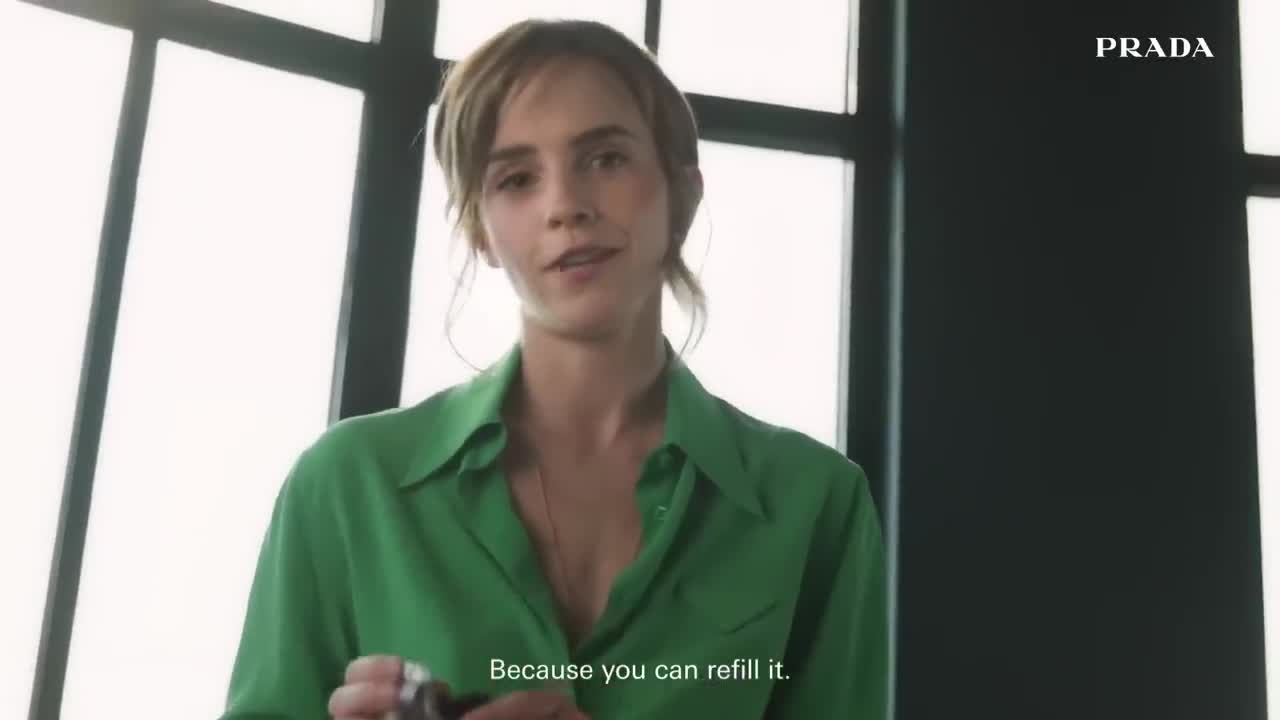} & \textbf{00:10} \quad A woman in a green top and yellow belt is standing by a window, holding a small bottle and a plate. The text \tabularnewline[4pt]
\includegraphics[width=0.30\textwidth]{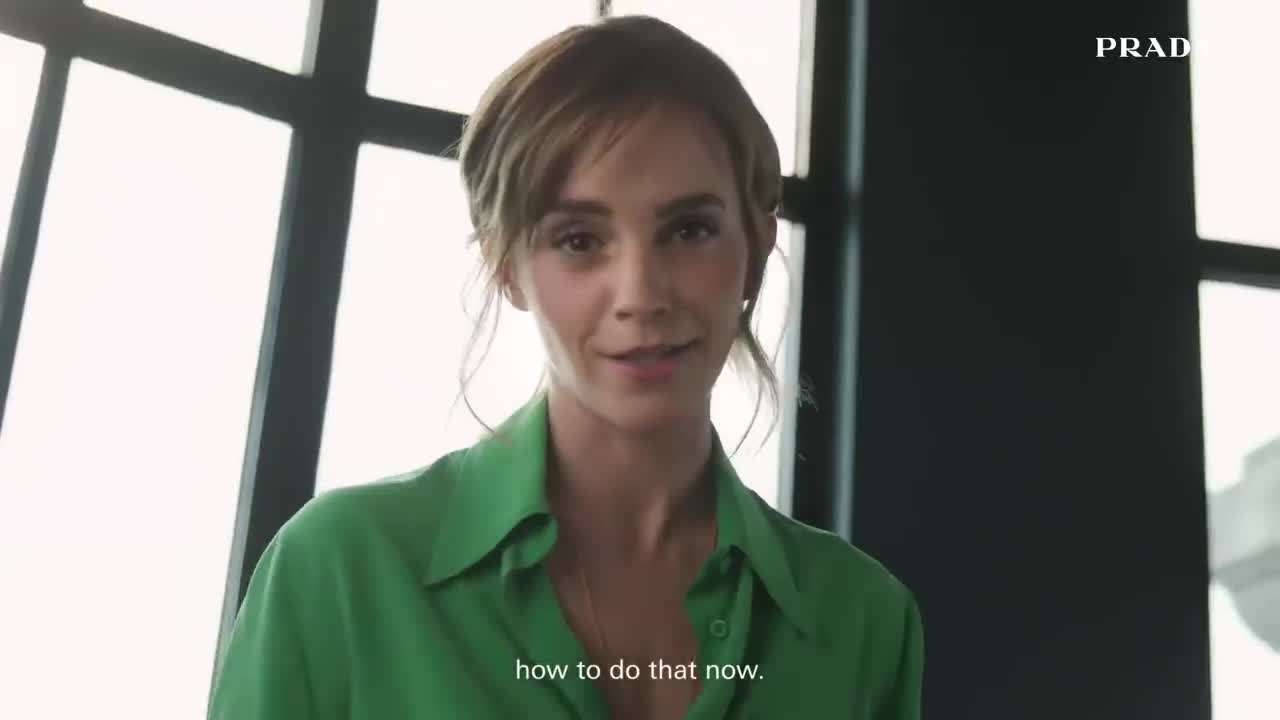} & \textbf{00:12} \quad ``PARADOX'' appears on screen. She is still holding the small bottle and plate as the scene continues. \tabularnewline[4pt]
\includegraphics[width=0.30\textwidth]{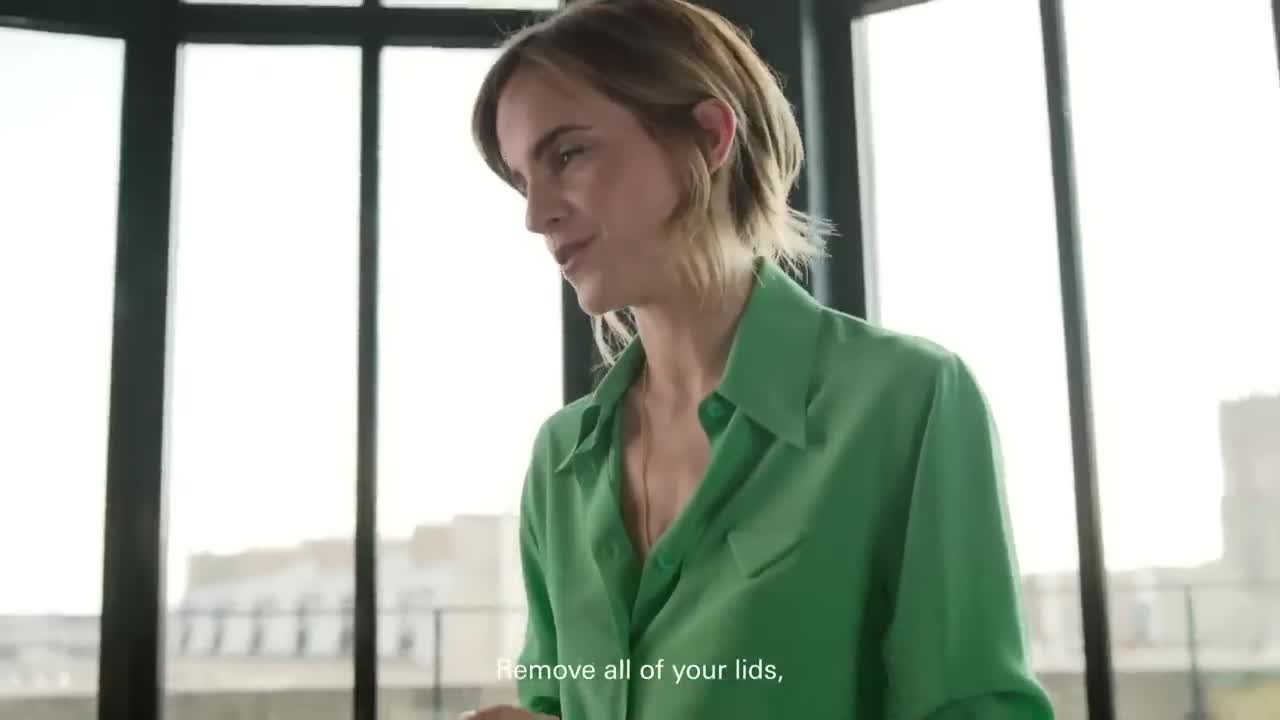} & \textbf{00:14} \quad The text ``PARADOX'' remains visible while she continues presenting the items. \tabularnewline[4pt]
\includegraphics[width=0.30\textwidth]{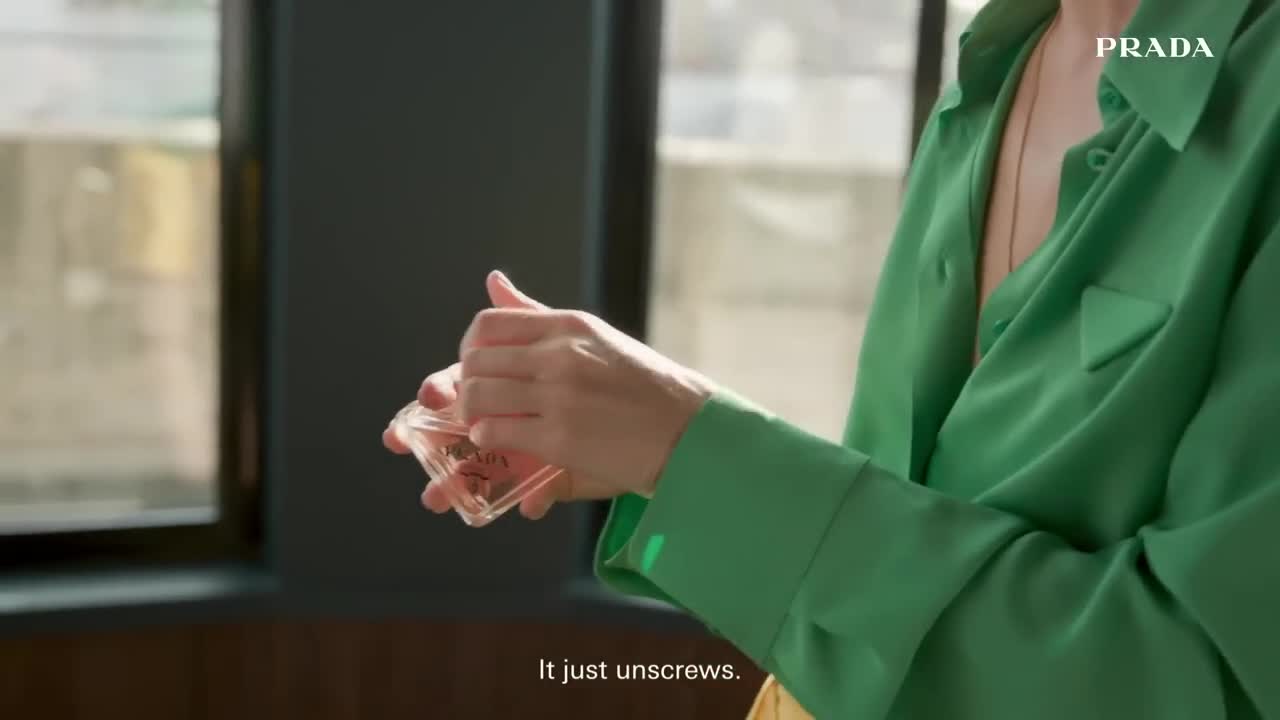} & \textbf{00:16} \quad She remains by the window with the bottle and plate as the shot continues. \\
\end{tabular}
\end{center}
\end{AIbox}
\end{minipage}
\end{figure*}

\clearpage

\begin{figure*}[p]
\centering
\begin{minipage}{0.92\textwidth}
\begin{AIbox}[width=\linewidth,top=6pt,bottom=4pt]{Cooking process with evolving observations}
\begin{center}
\begin{tabular}{@{}c@{\hspace{12pt}}p{0.56\textwidth}@{}}
\multirow{2}{*}{\sampleimage{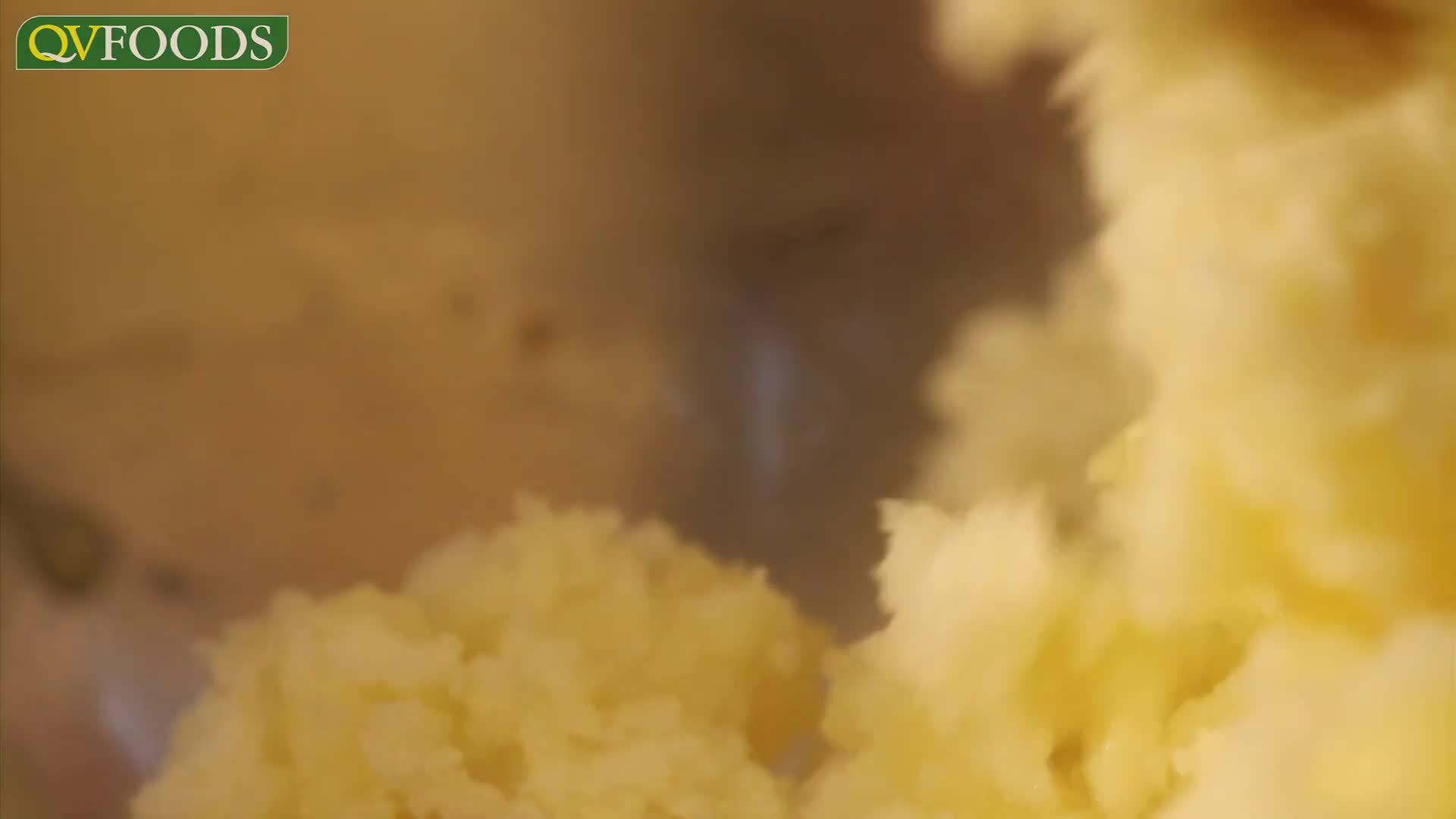}} & \samplehalfrow\textbf{Q (00:10):} That looks really buttery. \\
& \samplehalfrow\textbf{A:} Yeah, she's stirring a soft, grainy mixture in the bowl, and there are visible chunks of butter melting through it. \tabularnewline[4pt]
\multirow{2}{*}{\sampleimage{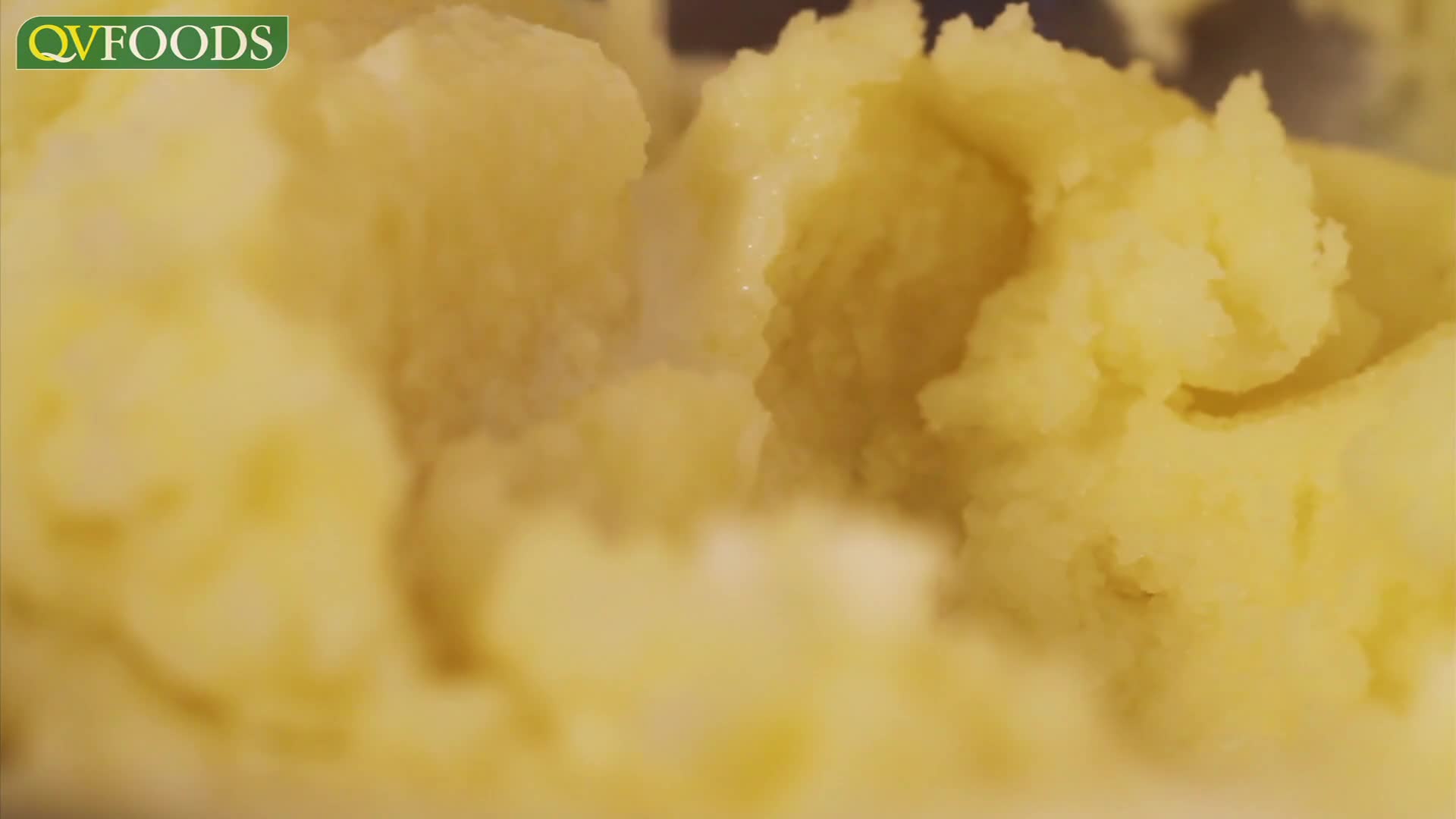}} & \samplehalfrow\textbf{Q (00:17):} It's looking smoother now. \\
& \samplehalfrow\textbf{A:} Yeah, the butter's worked in more and the mixture has turned into a softer, mashed texture. \tabularnewline[4pt]
\multirow{2}{*}{\sampleimage{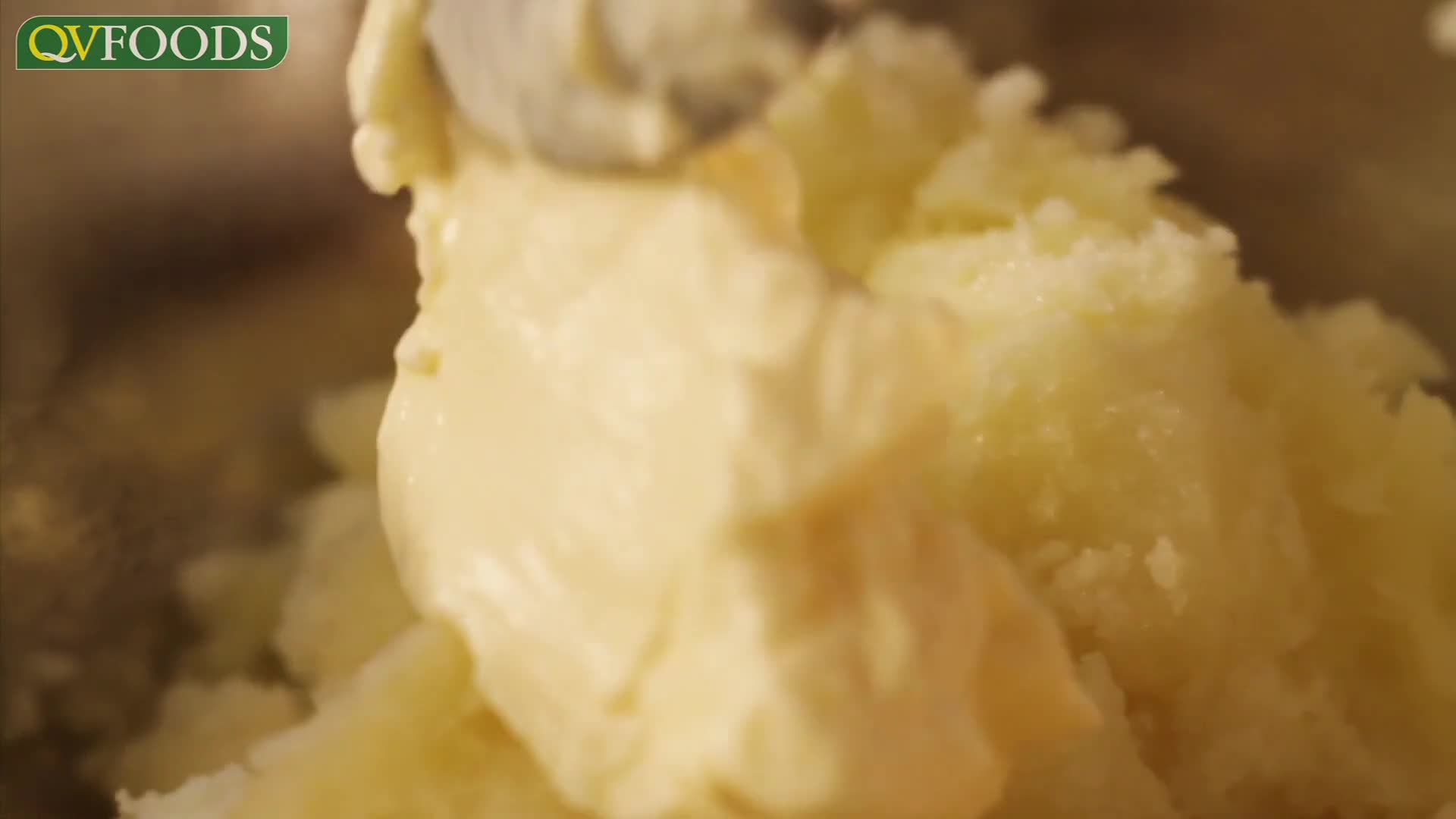}} & \samplehalfrow\textbf{Q (00:24):} Did she just add cream or milk there? \\
& \samplehalfrow\textbf{A:} Yes, she's pouring in a bit of milk from the white jug, then folding it through the mash. \tabularnewline[4pt]
\multirow{2}{*}{\sampleimage{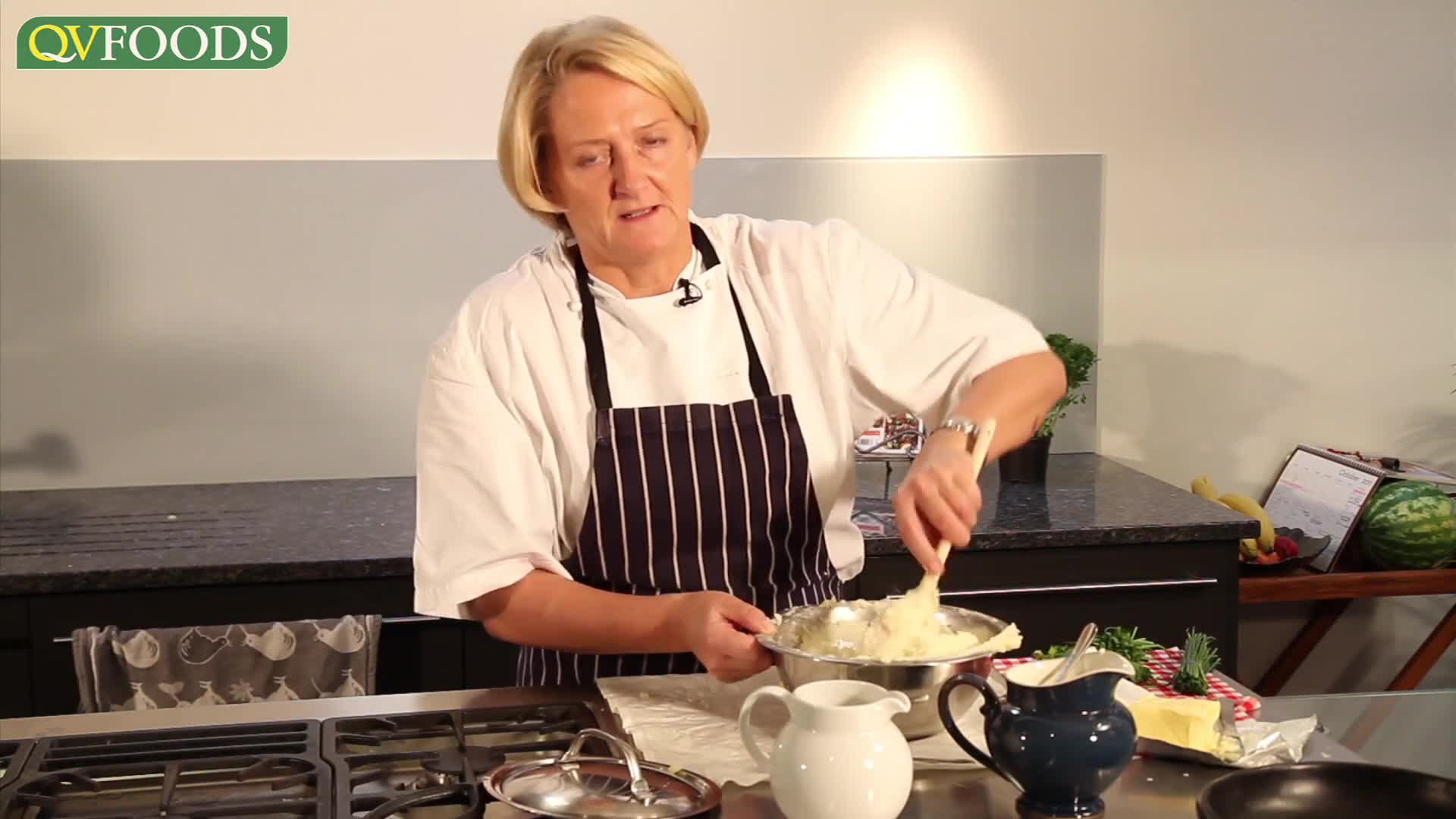}} & \samplehalfrow\textbf{Q (00:33):} She's adding something else now, right? \\
& \samplehalfrow\textbf{A:} Yes---she's adding a dollop of cream or soft cheese to the mash, making it look much silkier as she folds it in. \\
\end{tabular}
\end{center}
\end{AIbox}
\end{minipage}
\end{figure*}

\begin{figure*}[p]
\centering
\begin{minipage}{0.92\textwidth}
\begin{AIbox}[width=\linewidth,top=6pt,bottom=4pt]{Social interaction over time}
\begin{center}
\begin{tabular}{@{}c@{\hspace{12pt}}p{0.56\textwidth}@{}}
\multirow{2}{*}{\sampleimage{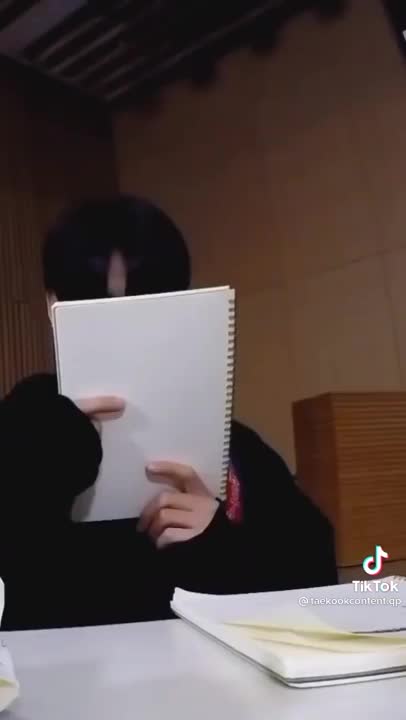}} & \samplehalfrow\textbf{Q (00:09):} He was really quiet, and now he's showing his notebook to someone. \\
& \samplehalfrow\textbf{A:} Yeah, he stops writing and holds the sketchbook out toward the person beside him, like he's asking for a look. \tabularnewline[4pt]
\multirow{2}{*}{\sampleimage{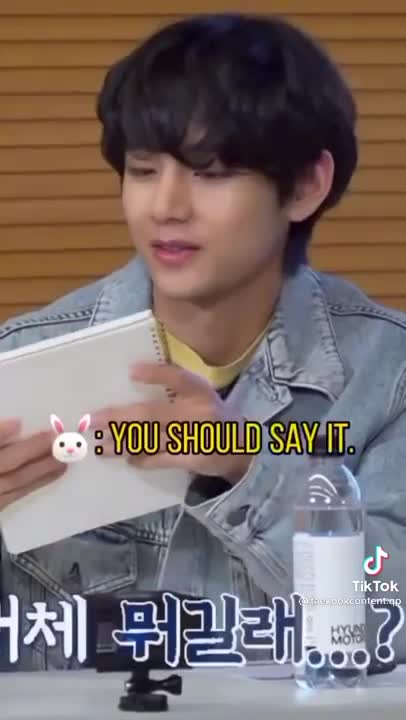}} & \samplehalfrow\textbf{Q (00:17):} Wait, did the other guy actually like what he wrote? \\
& \samplehalfrow\textbf{A:} Yeah, he looks impressed now and keeps asking why he didn't share it earlier. \tabularnewline[4pt]
\multirow{2}{*}{\sampleimage{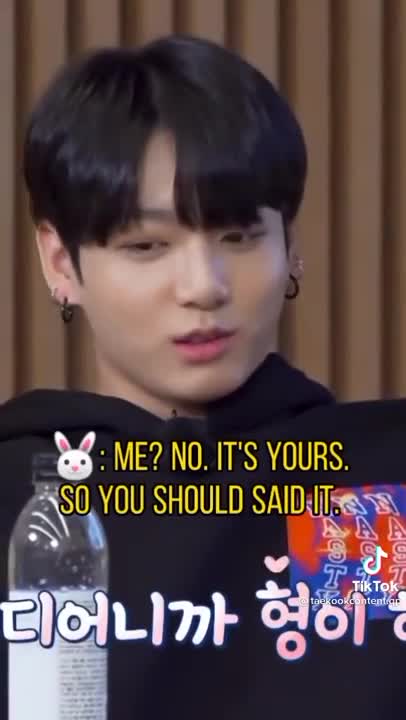}} & \samplehalfrow\textbf{Q (00:23):} Aw, now he's getting shy about it. \\
& \samplehalfrow\textbf{A:} Yeah, he's hiding behind the notebook while the other guy keeps insisting the idea is his and he should be the one to say it. \tabularnewline[4pt]
\multirow{2}{*}{\sampleimage{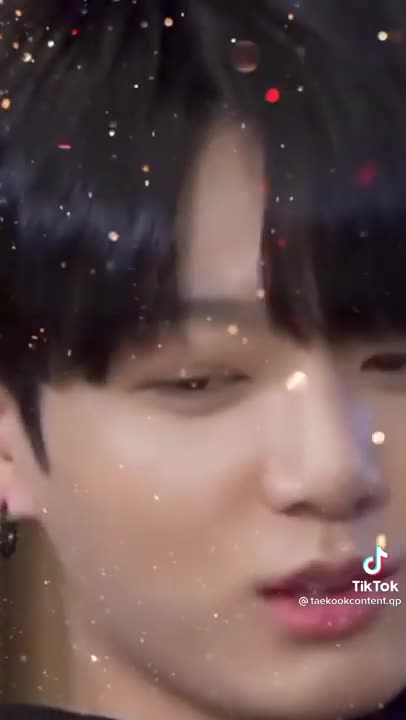}} & \samplehalfrow\textbf{Q (00:29):} He really won't take credit for it, huh? \\
& \samplehalfrow\textbf{A:} Yeah, he's still trying to hand the idea over, and the other guy keeps pushing it back to him. \\
\end{tabular}
\end{center}
\end{AIbox}
\end{minipage}
\end{figure*}

\begin{figure*}[p]
\centering
\begin{minipage}{0.92\textwidth}
\begin{AIbox}[width=\linewidth,top=6pt,bottom=4pt]{Tracking a rapidly changing montage}
\begin{center}
\begin{tabular}{@{}c@{\hspace{12pt}}p{0.56\textwidth}@{}}
\multirow{2}{*}{\sampleimage{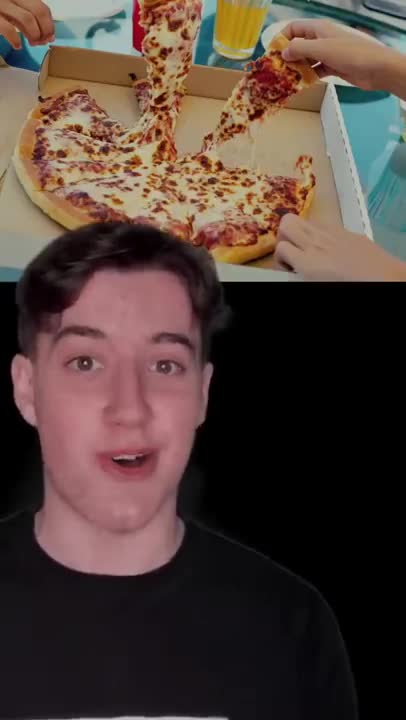}} & \samplehalfrow\textbf{Q (00:10):} That cut was abrupt---did it just switch scenes twice? \\
& \samplehalfrow\textbf{A:} Yeah, it jumped from the microphone setup to someone by a white car, then straight to a pizza close-up. \tabularnewline[4pt]
\multirow{2}{*}{\sampleimage{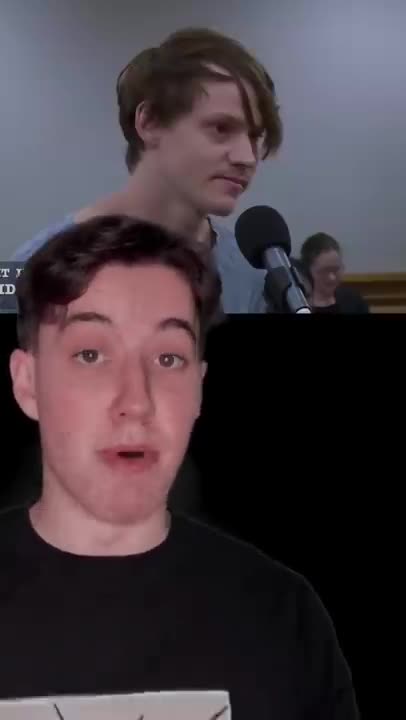}} & \samplehalfrow\textbf{Q (00:17):} It keeps cutting to random examples now---car, pizza, kid, police lights, even a red light. \\
& \samplehalfrow\textbf{A:} Yeah, it's turned into a rapid montage of examples while the two people by the microphone stay as the main setup. \tabularnewline[4pt]
\multirow{2}{*}{\sampleimage{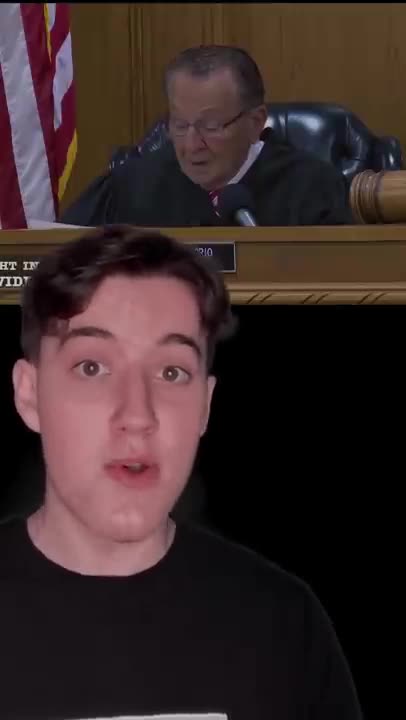}} & \samplehalfrow\textbf{Q (00:24):} Now it's connecting the montage back to the courtroom, right? \\
& \samplehalfrow\textbf{A:} Yeah, it cut from those example shots back to the person speaking at the microphone, and then wider courtroom views. \tabularnewline[4pt]
\multirow{2}{*}{\sampleimage{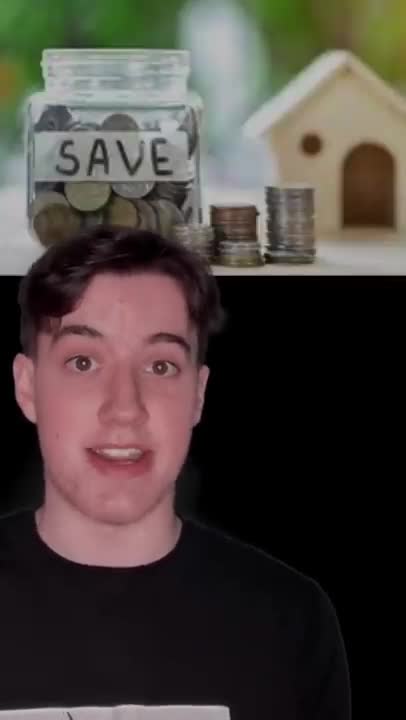}} & \samplehalfrow\textbf{Q (00:32):} So it's back to the montage again---now tickets, a wedding clip, and saving money? \\
& \samplehalfrow\textbf{A:} Yeah, it keeps bouncing from the courtroom testimony to these quick illustrative cutaways. \\
\end{tabular}
\end{center}
\end{AIbox}
\end{minipage}
\end{figure*}

\end{document}